\documentclass{article} 
\usepackage[T1]{fontenc}
\usepackage[preprint]{colm2026_conference}

\usepackage{microtype}
\usepackage{hyperref}
\usepackage{url}
\usepackage{booktabs}
\usepackage{graphicx}
\usepackage{amsmath}
\usepackage{amssymb}
\usepackage{xspace}
\usepackage{xcolor}
\usepackage{enumitem}
\usepackage{titlesec}
\titlespacing*{\section}{0pt}{8pt plus 2pt minus 2pt}{4pt plus 1pt minus 1pt}
\titlespacing*{\subsection}{0pt}{6pt plus 2pt minus 2pt}{3pt plus 1pt minus 1pt}
\usepackage{subcaption}
\usepackage[ruled,vlined]{algorithm2e}
\usepackage{setspace}
\usepackage{tikz}
\usetikzlibrary{arrows.meta,positioning}
\usepackage{listings}
\newcommand{\opreflect}{\textsc{Reflect}\xspace}
\newcommand{\opcodegen}{\textsc{CodeGen}\xspace}
\newcommand{\opexec}{\textsc{Execute}\xspace}
\newcommand{\oppatch}{\textsc{Patch}\xspace}
\newcommand{\opplan}{\textsc{Plan}\xspace}

\usepackage{lineno}

\definecolor{darkblue}{rgb}{0, 0, 0.5}
\hypersetup{colorlinks=true, citecolor=darkblue,linkcolor=darkblue, urlcolor=darkblue}

\title{Evolving Programmatic Skill Networks}

\author{
\textbf{Haochen Shi}$^{1,2}$ \quad
\textbf{Xingdi Yuan}$^{3}$\thanks{Equal advising} \quad
\textbf{Bang Liu}$^{1,2,4}$\footnotemark[1]
\\[0.4em]
$^{1}$ DIRO \& Institut Courtois, Université de Montréal \quad
$^{2}$ Mila -- Québec AI Institute \\
$^{3}$ Microsoft Research
\quad
$^{4}$ Canada CIFAR AI Chair
\\[0.4em]
\texttt{haochen.shi@umontreal.ca} \quad
\texttt{eric.yuan@microsoft.com} \quad
\texttt{bang.liu@umontreal.ca}
}

\begin{document}

\ifcolmsubmission
\linenumbers
\fi

\maketitle

\begin{abstract}
We study continual skill acquisition in open-ended embodied environments where an agent must construct, refine, and reuse an expanding library of executable skills. We introduce the Programmatic Skill Network (PSN), a framework in which skills are executable symbolic programs forming a compositional network that evolves through experience.
PSN defines three core mechanisms instantiated via large language models: (1)~\opreflect for structured fault localization over skill compositions, (2)~progressive optimization with maturity-aware update gating that stabilizes reliable skills while maintaining plasticity for uncertain ones, and (3)~canonical structural refactoring under rollback validation that maintains network compactness. 
We further show that PSN's learning dynamics exhibit structural parallels to neural network training.
Experiments on MineDojo and Crafter demonstrate robust skill reuse, rapid adaptation, and strong generalization across open-ended task distributions.\footnote{Code: \href{https://github.com/evolving-skill-networks/psn}{github.com/evolving-skill-networks/psn}; Webpage: \href{https://evolving-skill-networks.github.io}{evolving-skill-networks.github.io}.}
\end{abstract}

\section{Introduction}

Embodied agents operating in open-ended environments must continually acquire, refine, and reuse a growing repertoire of skills. Existing approaches~\citep{wang2024voyager,yao2023react} suffer from two limitations: (1) skills are typically represented as flat libraries or static graphs lacking principled mechanisms for continual improvement, and (2) agents lack unified frameworks for assigning credit over hierarchical skill compositions, repairing symbolic programs, and reorganizing structure as new tasks arise.

We introduce the Programmatic Skill Network (PSN), a framework for continually evolving skill libraries. In a PSN, each skill is a symbolic program (e.g., in JavaScript for Minecraft, Python for Crafter) with explicit control flow, parameters, and preconditions that specify applicability and effects. Skills invoke each other through dependency links, forming a directed graph that grows and reorganizes as the agent learns. While recent work has explored programmatic skill representations for agents \citep{wang2024codeact,stengel-eskin2024regal,wang2025asi}, PSN uniquely maintains an explicit computational graph of executable programs that supports trace-based credit assignment, maturity-aware stabilization, and principled structural refactoring.

\begin{figure*}[t!]
\centering
\includegraphics[width=\textwidth]{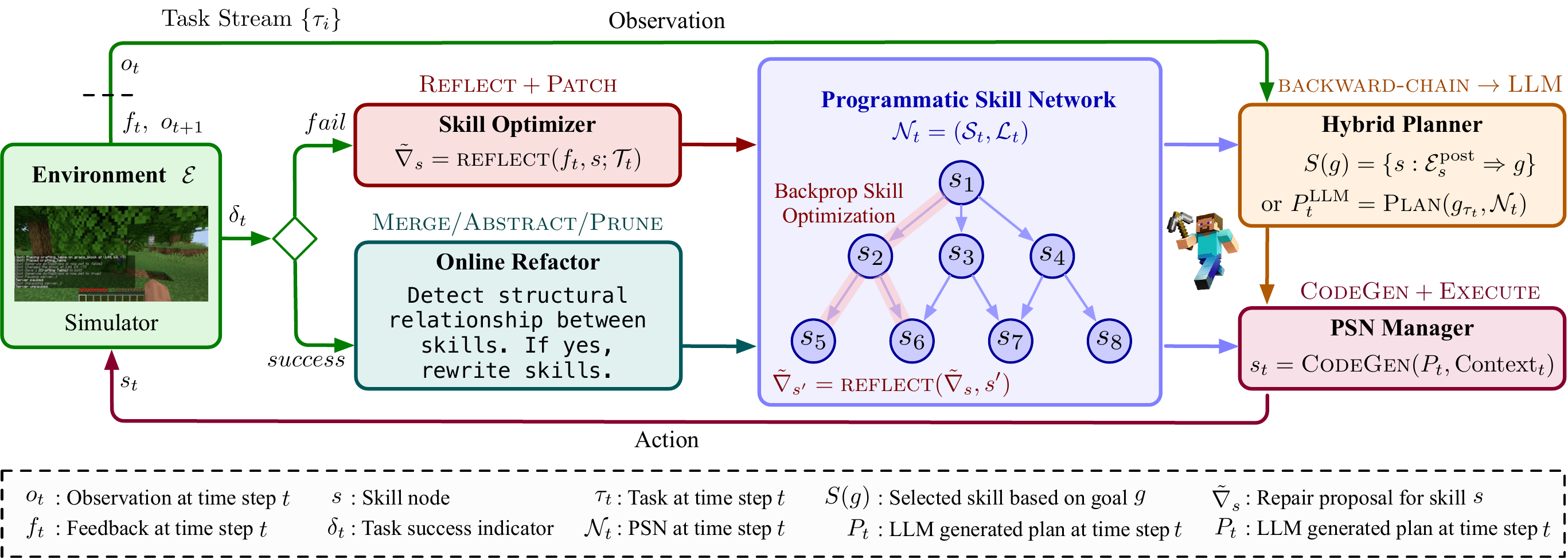}
\caption{The Programmatic Skill Network (PSN) framework. The agent maintains a skill network $\mathcal{N}_t$ where the \textit{hybrid planner} selects or synthesizes skills; the \textit{PSN manager} executes them. On failure, the \textit{skill optimizer} performs trace-based credit assignment; on success, the \textit{online refactor} restructures the network. This induces learning dynamics analogous to neural network training: fault localization as backpropagation, maturity gating as learning rate scheduling, and refactoring as architecture search.}
\label{fig:framework}
\end{figure*}

The framework structures continual learning through three components: a \emph{network-aware planner} that prioritizes skill reuse via backward-chaining, a \emph{fault localization mechanism} (\textsc{Reflect}) that assigns credit over skill compositions by analyzing execution traces, and a \emph{refactor module} that reorganizes network structure. These components are instantiated using LLMs for program synthesis, but the continual learning behavior emerges from the architectural scaffolding rather than the LLM itself. Figure~\ref{fig:framework} provides an overview of the PSN framework, illustrating the agent--environment interaction under a curriculum task stream (left) and the internal evolution of the programmatic skill network through planning, repairing, and structural refactoring (right).

A key insight is that PSN's learning dynamics exhibit structural parallels to neural network training. Fault localization over skill compositions resembles backpropagation through computational graphs~\citep{rumelhart1986backprop}; maturity-based update gating induces stability-plasticity tradeoffs analogous to layer freezing and learning rate scheduling~\citep{howard2018ulmfit,yosinski2014transfer,rusu2016progressive}; and structural refactoring performs a form of symbolic neural architecture search~\citep{zoph2017nas,han2016deepcompression,tan2019efficientnet}. These parallels suggest that principles of neural network optimization extend to programmatic learning systems.

The contributions of this work are threefold:\\
\noindent$\bullet$ \textbf{Programmatic Skill Networks.} We introduce a framework for continual skill learning in which skills are executable symbolic programs with explicit control flow, parameters, and pre/postconditions, forming a compositional network through invocation links and yielding an inspectable computational graph that grows and reorganizes as the agent learns.\\
\noindent$\bullet$ \textbf{PSN learning mechanisms.} We develop three complementary mechanisms for continual skill improvement: (1)~\opreflect for fault localization; (2)~maturity-aware update gating for stabilizing reliable skills while maintaining plasticity for uncertain ones; and (3)~canonical structural refactoring with rollback validation for eliminating redundancy while preserving performance.\\
\noindent$\bullet$ \textbf{An optimization perspective.} We show that PSN's architectural design induces learning dynamics with structural parallels to neural network training, suggesting general principles for continual learning across representational paradigms.

\section{Method}
\label{section:method}

\textbf{Problem setup.} We consider an embodied agent acting in a partially observable Markov decision process (POMDP)~\citep{kaelbling1998pomdp}. 
The agent receives a stream of open-ended tasks $T = \{\tau_1, \tau_2, \ldots\}$,
each specified in natural language and associated with a goal predicate $g_\tau : \mathcal{S} \to \{0, 1\}$, where $\mathcal{S}$ denotes the state space. Tasks arrive sequentially and may vary in difficulty, horizon length, and compositional structure. The agent must continually acquire, refine, and reorganize reusable skills to solve future tasks by leveraging past experience.

We present an online framework for continually constructing, optimizing, and refactoring a Programmatic Skill Network. It evolves through a recurrent loop that couples symbolic planning, execution, failure-driven repair, and success-driven structural refactoring. We first define the core objects and operators that constitute the network, then describe the planning and learning mechanisms.

\subsection{Programmatic Skill Networks (PSN)}
\label{subsec:psn}

A skill $s = (\mathcal{C}_s, \mathcal{P}_s, \mathcal{E}_s, \textsc{Children}(s))$ is a symbolic program where $\mathcal{C}_s$ denotes control flow, $\mathcal{P}_s$ parameters, $\mathcal{E}_s = (\mathcal{E}_s^{\text{pre}}, \mathcal{E}_s^{\text{post}})$ preconditions/postconditions, and $\textsc{Children}(s)$ invoked subskills. This precondition-effect structure is analogous to programmatic laws in symbolic world modeling \citep{khan2025one}. The agent maintains a directed network $\mathcal{N}_t = (\mathcal{S}_t, \mathcal{L}_t)$ where nodes $\mathcal{S}_t$ are skills and edges $\mathcal{L}_t$ represent invocations.

Executing skill $s$ yields $(f_s, \delta_s)$ where $\delta_s \in \{0, 1\}$ indicates success and $f_s$ aggregates feedback from the environment. The system records a finite invocation trace $\mathcal{T}$. 
Given feedback $f_s$, \opreflect computes repair proposal $\tilde{\nabla}_s$ identifying faulty control flow, preconditions, parameters, or subskills. 
For invoked subskills $s' \in \text{Children}(s)$, responsibility propagates as
\vspace{-0.3em}
\begin{equation}
\tilde{\nabla}_{s'} = \opreflect(\tilde{\nabla}_s, s'),
\end{equation}
\vspace{-0.1em}
yielding finite credit assignment over executed subgraphs.

Each skill maintains scalar value $V(s) = \hat{p}_s - u_s$ where $\hat{p}_s$ is success rate with Laplace smoothing and $u_s$ is an uncertainty term that decreases as more executions are observed. This value summarizes long-term skill reliability and serves a dual role: guiding skill selection during planning and modulating update frequency during optimization.


Beyond behavioral repair, the PSN evolves through structure-level rewrites such as merging redundant skills, abstracting shared routines, pruning irrelevant branches, and rewiring invocation links. These operations are treated as discrete architecture updates and are validated through rollback-based safety checks (Section~\ref{subsec:refactor}).

\paragraph{LLM implementation.} In our implementation, operators such as \opreflect are instantiated via prompted LLMs. The framework defines information flow \textit{structure} (e.g., what information is available, output formats, update timing) while LLMs provide the generative capacity to synthesize, diagnose, and repair programs within this structure.
Critically, the learning dynamics we observe (Section~\ref{sec:optimization}) emerge from the \textit{architectural choices} of PSN (e.g., the compositional network structure, the execution trace-based credit assignment, the maturity-gated updates, and the canonical refactor operations) rather than from the internal mechanisms of the LLM. This separation allows the framework to be instantiated with different code generation backends while preserving its continual learning properties.


\subsection{Network-Aware Hybrid Planner}
\label{subsec:planner}
The planner prioritizes reuse of the existing PSN via symbolic backward-chaining before invoking LLM-based forward planning. Each skill $s$ is treated as an operator with preconditions $\mathcal{E}_s^{\text{pre}}$ and postconditions $\mathcal{E}_s^{\text{post}}$. Starting from the goal predicate, the planner selects skills whose postconditions satisfy current subgoals:
\begin{equation}
S(g) = \{s : \mathcal{E}_s^{\text{post}} \Rightarrow g\},
\end{equation}
\vspace{-0.1em}
and recursively expands unmet preconditions. 
When multiple skills satisfy a subgoal, ties are broken by $V(s)$, favoring skills with higher empirical reliability.
Skill selection uses Boltzmann exploration~\cite{sutton1998reinforcement} over the value function $V(s)$, balancing exploitation of reliable skills with exploration of uncertain ones.
If no skill can reduce a subgoal, the planner invokes an LLM-based forward planner $P_t^{\text{LLM}} = \opplan(g_{\tau_t}, \mathcal{N}_t)$. Successful plans are distilled into new skills via the execution pipeline described next.

\subsection{Execution and Trace Construction}
\label{subsec:manager}
Given a plan $P_t = [s_1, \ldots, s_k]$, the PSN manager synthesizes a candidate skill
\begin{equation}
s_t = \opcodegen(P_t, \text{Context}_t),
\end{equation}
\vspace{-0.1em}
where $\text{Context}_t$ includes the task description, current network $\mathcal{N}_t$, and execution history. The synthesized skill defines control flow $\mathcal{C}_{s_t}$, parameters $\mathcal{P}_{s_t}$, and pre/postconditions $\mathcal{E}_{s_t}$, and is inserted into the PSN with invocation links to its children.
Executing $s_t$ produces a skill execution trace:
\begin{equation}
\opexec(s_t) \rightarrow (f_t, \delta_t, \mathcal{T}_t),
\end{equation}
\vspace{-0.1em}
where $\delta_t \in \{0,1\}$ indicates task success, $f_t$ aggregates environment feedback and critic signals, and the trace $\mathcal{T}_t$ records each invoked skill as a tuple $\langle s, \sigma^{\text{pre}}, \sigma^{\text{post}}, \text{status} \rangle$ with symbolic state snapshots $\sigma$. The trace serves as supervision for both optimization and refactoring. Preconditions and postconditions are incrementally calibrated from observed success/fail states and empirical transitions.

\subsection{Skill Optimization via Trace-Based Credit Assignment}
\label{subsec:skill_opt}

When execution fails (i.e., $\delta_t = 0$), the skill optimizer performs localized behavioral repair via structured fault localization. Unlike approaches that discover world dynamics in natural language \citep{sun-etal-2024-enhancing-agent} or learn function libraries offline \citep{stengel-eskin2024regal}, PSN performs online, trace-based credit assignment over executable skill compositions. Given feedback $f_t$ and trace $\mathcal{T}_t$, the \textsc{Reflect} operator computes a repair proposal for each executed skill:
\begin{equation}
\tilde{\nabla}_s = \opreflect(f_t, s; \mathcal{T}_t),
\end{equation}
\vspace{-0.1em}
identifying faulty control flow, violated preconditions, misaligned parameters, or incorrect subskill effects. 
Concretely, PSN separates \emph{credit assignment} from \emph{code modification} through a two-phase process: failure signals are first propagated \emph{top-down} along the executed skill invocation trace to decompose responsibility across composite skills and their subskills (symbolic differentiation), after which localized symbolic edits are applied \emph{bottom-up} to individual skills in a dependency-respecting order (gradient application).
Proposals propagate in reverse execution order along the invocation trace; skills not in $\mathcal{T}_t$ receive no updates. Each affected skill is updated via $s \leftarrow \oppatch(s, \tilde{\nabla}_s)$. 
The complete two-phase optimization procedure of the skill optimizer, including the top-down symbolic differentiation and bottom-up gradient application are described in Appendix~\ref{app:two_phase_optimizer}.

To stabilize learning, updates are constrained by a rolling buffer of the 5 most recent repair proposals, preventing contradictory edits. Update frequency is further modulated by skill maturity:
\begin{equation}
P(\text{update } s) = (1-\epsilon) \cdot \sigma(\gamma(0.6 - V(s))) + \epsilon,
\end{equation}
\vspace{-0.1em}
The constant $0.6$ serves as a soft maturity pivot rather than a bound on $V(s)$: it marks the inflection point at which a skill is considered sufficiently reliable to gradually reduce update frequency, while still allowing occasional repairs under compositional failures. $\sigma$ is the sigmoid function, $\gamma = 5.0$ controls threshold sharpness, and $\epsilon = 0.1$ ensures minimum update probability. Mature skills ($V(s) \approx 1$) stabilize with low update probability, while immature skills remain plastic.

\subsection{Online Structural Refactoring}
\label{subsec:refactor}

The online skill refactor controls structural growth via semantics-preserving refactorings, applying architecture-level rewrites that increase skill reuse and maintain network compactness. While code refactoring has been used to discover generalizable abstractions offline \citep{stengel-eskin2024regal}, PSN performs online refactoring that adapts to errors and redundancies emerging during continual learning. While the skill optimizer repairs individual skill programs, refactor operates at the network level, targeting redundancy and missed abstractions that emerge over continual learning.
\paragraph{Canonical refactor cases.} 
We restrict refactor to five structural relationships: (i) \emph{Parametric coverage}: one skill is a strict specialization of another admitting parameterized generalization. (ii) \emph{Behavioral coverage}: a composite skill reimplements existing functionality. (iii) \emph{Sibling specializations}: multiple skills suggest a missing abstraction. (iv) \emph{Common subskill extraction}: multiple skills share identical sub-operations. (v) \emph{Duplication}: two skills are functionally equivalent. Each admits a fixed rewrite rule; visual illustrations are provided in Appendix~\ref{app:refactor-casebook}.

\paragraph{Candidate discovery and rewrites.} 
Given a successfully executed skill $s_t$, refactor operates on a restricted candidate set: parents and children of $s_t$, plus top-5 semantically related skills by embedding similarity. For each detected relationship, deterministic rewrites are applied (wrapper conversion, call substitution, abstract skill synthesis, shared subskill extraction, or canonical merging). Refactor does not introduce new behavioral logic, it only reorganizes existing programs and invocation links.

\paragraph{Safety via rollback validation.} 
All refactor proposals are tentative. Given a refactored candidate network $\mathcal{N}'_t$, the system evaluates short-horizon performance on a sliding window of 3 recent tasks involving affected skills. If the task success rate drops by more than 20\%, the refactor is reverted using logged inverse operations. Empirical rollback rates and refactoring case distributions are reported in Appendix~\ref{app:rollback_stats} and Appendix~\ref{app:refactor_distribution}.

\begin{table*}[t!]
\small
\centering
\resizebox{\textwidth}{!}{%
\begin{tabular}{l|lllll}
\toprule
\textbf{Method} & \textbf{Wooden Tool} & \textbf{Stone Tool} & \textbf{Iron Tool} & \textbf{Diamond Tool} & \textbf{Obsidian}\\
\midrule
ReAct & N/A (0/3) & N/A (0/3) & N/A (0/3) & N/A (0/3) & -- \\
Reflexion & N/A (0/3) & N/A (0/3) & N/A (0/3) & N/A (0/3) & -- \\
AutoGPT & 92 $\pm$ 72 (3/3) & 94 $\pm$ 72 (3/3) & 135 $\pm$ 103 (3/3) & N/A (0/3) & -- \\
Voyager & 6 $\pm$ 2 (3/3) & 11 $\pm$ 2 (3/3) & 21 $\pm$ 7 (3/3) & 102 (1/3) & -- \\
Voyager* & 6 $\pm$ 3 (6/6) & 13 $\pm$ 4 (6/6) & 34 $\pm$ 10 (6/6) & 99 $\pm$ 36 (2/6) & N/A (0/6)\\
ADAM* & 21 $\pm$ 4 (3/3) & 36 $\pm$ 5 (3/3) & 52 $\pm$ 15 (3/3) & 74 $\pm$ 23 (3/3)& N/A (0/3) \\
PSN w/o Optimizer &\textbf{5 $\pm$ 2 (3/3)} & 12 $\pm$ 2 (3/3) & 25 $\pm$ 4 (3/3) & N/A (0/3) & N/A (0/3)\\
PSN (Ours) & \textbf{5 $\pm$ 2 (6/6)} & \textbf{10 $\pm$ 3 (6/6)} & \textbf{22 $\pm$ 5 (6/6)} & \textbf{35 $\pm$ 16 (6/6)} & \textbf{77 (1/6)} \\
\midrule
Voyager$^\dagger$ & 9 $\pm$ 4 (6/6) & 24 $\pm$ 23 (6/6) & 60 $\pm$ 41 (4/6) & 94 (1/6) & N/A (0/6) \\
ADAM$^\dagger$ & 43  $\pm$ 8 (2/3) & 65 (1/3) & N/A (0/3) & N/A (0/3) & N/A (0/3) \\
PSN$^\dagger$ (Ours) & \textbf{7 $\pm$ 3 (6/6)} & \textbf{18 $\pm$ 7 (6/6)} & \textbf{38 $\pm$ 9 (6/6)} & \textbf{49 $\pm$ 18 (6/6)} & N/A (0/6) \\
\bottomrule
\end{tabular}%
}
\caption{Tech tree mastery on Minecraft.
We report the mean/std iterations an agent uses to unlock an item. PSN and our Voyager reproductions (* and $\dagger$) are averaged over six runs; the remaining baselines over three runs.
For example, PSN successfully unlocks the diamond tool in all six runs, on average using 35 iterations; while Voyager~\cite{wang2024voyager} succeeds in one of three runs using 102 iterations. * indicates results obtained using the open-sourced code with \texttt{GPT-5-mini} (same as ours). $\dagger$ indicates results obtained using \texttt{Qwen3-Coder-Next} (3B active parameters, 80B total open-weight, MoE). N/A represents the failure to unlock an item across all runs. -- represents unreported previous result.}
\label{tab:minecraft_main}
\end{table*}

\section{An Optimization Perspective on PSN}
\label{sec:optimization}

Having presented PSN's concrete mechanisms (Section~\ref{section:method}), we can observe that the system's learning dynamics exhibit structural parallels to neural network training. While other neuro-symbolic systems embed symbolic rules inside differentiable models \citep{garcez2019neural,manhaeve2018deepproblog} or use gradient-free skill-based routing \citep{chen2025symbolic}, PSN embeds learning dynamics inside symbolic programs. This interpretive lens clarifies how PSN's architectural choices collectively induce coherent continual learning behavior, independent of the LLM backend.

\paragraph{Implicit structure-behavior trade-off.}
Let $\mathcal{N} = (\mathcal{S}, \mathcal{L})$ denote the current PSN. The system's behavior can be viewed as implicitly optimizing a composite objective:
\begin{equation}
J(\mathcal{N}) = \mathcal{R}_{\text{task}} + \mathcal{R}_{\text{reliab}} + \mathcal{R}_{\text{struct}} + \mathcal{R}_{\text{cons}},
\end{equation}
balancing \textbf{task} success, skill \textbf{reliab}ility, \textbf{struct}ural compactness, and semantic \textbf{cons}istency. While never explicitly optimized, each PSN module performs localized improvements to different components of $J(\mathcal{N})$.

\paragraph{Operator-objective correspondence.}
\opreflect acts as \emph{symbolic differentiation}: when a task fails, it identifies which control-flow branches, preconditions, parameters, and subskill compositions contributed to the error, producing structured repair proposals that reduce $\mathcal{R}_{\text{task}}$ and $\mathcal{R}_{\text{cons}}$. Like backpropagation, credit is assigned only along the executed path, with non-executed skills receiving no updates.
This selective credit assignment avoids the noise of updating uninvolved skills, mirroring how gradients flow only through activated paths in neural nets.
A chain-rule view of this credit assignment makes a falsifiable prediction: a noisier repair signal must be propagated through more compositional layers to localize the same fault, so a weaker LLM should require deeper REFLECT chains. We observe exactly this (5.0 vs 2.7 skills per multi-skill repair; Section~\ref{sec:cross_llm}).
A quantitative analysis of credit assignment patterns, error mode distributions, and attribution coherence is provided in Appendix~\ref{app:credit_assignment}.
Maturity-aware gating functions as \emph{adaptive learning rates}: mature skills with high $V(s)$ receive infrequent updates (analogous to freezing converged layers), while immature skills remain plastic, reducing $\mathcal{R}_{\text{reliab}}$ by preventing catastrophic forgetting.
Refactor performs \emph{symbolic neural architecture search}: merging redundant skills, extracting reusable abstractions, and pruning unnecessary branches to reduce $\mathcal{R}_{\text{struct}}$. Rollback-based validation functions as a symbolic trust region.
The depth of this search is model-dependent: it converges to a hub-and-spoke graph with reuse ratio $\mathcal{R}_{\text{struct}}\!\approx\!0.4$ for the stronger model but a flatter $0.15\pm0.07$ for the weaker one (Section~\ref{sec:cross_llm}), the structural counterpart of the deeper repair chains above.

\paragraph{Multi-scale learning dynamics.}
PSN learning unfolds across three coupled timescales: (1)~\emph{Fast}: fault localization performs frequent behavioral repair at every execution. (2)~\emph{Intermediate}: maturity-based stabilization progressively freezes reliable skills over 10--50 executions. (3)~\emph{Slow}: structural refactor reorganizes stabilized behaviors every 5--10 successful executions. This yields a coherent dynamic: optimize behavior locally and rapidly, stabilize reliable skills over time, and restructure only after behaviors have converged.

\paragraph{Scope of the analogy.}
The neural network analogy is \textbf{partial}. PSN operates over discrete symbolic programs rather than continuous parameters, produces structured edit proposals rather than numeric derivatives, and relies on binary success/failure signals rather than differentiable losses. Nevertheless, it reveals that stability-plasticity tradeoffs, compositional credit assignment, and architecture search emerge as general principles when learning structured representations. This suggests that insights from neural network optimization may inform symbolic learning systems, and vice versa. An empirical decomposition of $J(\mathcal{N})$ into its four components across training is provided in Appendix~\ref{app:jn_decomposition}.

\section{Experiments and Analysis}
\label{sec:experiments}

We evaluate Programmatic Skill Networks (PSN) on two complementary embodied benchmarks: \textbf{MineDojo}~\citep{fan2022minedojo}, which supports long-horizon open-ended Minecraft tasks with rich action spaces and diverse goal specifications, and \textbf{Crafter}~\citep{hafner2022crafter}, a lightweight survival environment with a structured technology progression that stresses continual learning and compositional reuse. 
Across both environments, we evaluate (i) end-task performance, (ii) continual learning dynamics (learning/forgetting), (iii) compositional generalization, and (iv) network structural properties (growth, reuse, redundancy) induced by refactor and maturity-aware optimization.


\subsection{Experimental Setup}
\label{subsec:setup}
We leverage OpenAI’s \texttt{gpt-5-mini-2025-08-07} for all operators across both environments. To assess whether PSN’s learning dynamics are robust across model families, we additionally run all Minecraft experiments with \texttt{Qwen3-Coder-Next}~\citep{cao2026qwen3}, an open-weight model with 3B active parameters (80B total), representing a fundamentally different model family from GPT-5-mini and orders of magnitude smaller in active model size.
The Minecraft simulator is built on top of MineDojo and leverages Mineflayer JavaScript APIs for motor controls~\citep{mineflayer_github}. For the Crafter environment, we implemented a Mineflayer-like Python API system for the control of the Crafter bot.
PSN operators (e.g., \opcodegen and \opreflect) are instantiated by prompted LLMs (see Appendix~\ref{app:prompt_templates} for example prompts).
The continual learning dynamics we measure arise from PSN's architectural scaffolding: trace-based credit assignment, maturity-aware update gating, and canonical refactor operations, rather than prompt-level tricks. We verify this attribution directly via a cross-LLM ablation of inline knowledge scaffolding (Appendix~\ref{app:scaffolding_ablation}), showing that on GPT-5-mini the inline knowledge corrigendum contributes only $+2$pp to diagnostic correctness on Phase~1 reflection cases, versus $+33$pp on Qwen3.

We compare PSN against representative LLM-agent baselines and ablations.
\textbf{ReAct}~\citep{yao2023react}, a prompting-based agent that interleaves reasoning and action without persistent structured skills.
\textbf{Reflexion}~\citep{shinn2023reflexion}, an agent self-reflects over failures but does not maintain a compositional programmatic skill network.
\textbf{AutoGPT}~\citep{autogpt}, a planning-centric agent that decomposes tasks into multi-step plans and executes generated code or action sequences autonomously. It maintains a short-term memory of past actions and observations, but treats generated plans and code fragments as ephemeral artifacts rather than persistent, reusable skills. 
\textbf{Voyager}~\citep{wang2024voyager}, an agent that maintains a flat skill library and retrieves skills via similarity, without trace-based symbolic credit assignment and canonical structural refactor as in PSN.
\textbf{ADAM}~\citep{zhang2024adam}, an agent that discovers causal precondition-effect mappings for a fixed set of 47 hand-authored JavaScript skills via LLM-guided inference and intervention-based verification (do-calculus), then uses the learned causal graph to guide LLM planning. Unlike PSN, ADAM does not generate or optimize skill code; its learning is restricted to discovering input-output relationships of predefined skills. We run ADAM's original code with the same LLMs as PSN for controlled comparison.
\textbf{ODYSSEY}~\citep{liu2024odyssey}, an agent that pre-builds 183 hand-authored JavaScript skills and reduces the LLM's role to retrieval-based skill selection rather than code generation; we compare architecturally in Appendix~\ref{app:odyssey_comparison}.
Other Minecraft LLM agents (e.g., DEPS~\citep{wang2023deps}, JARVIS-1~\citep{wang2024jarvis1}, Optimus-1~\citep{li2024optimus1}, RL-GPT~\citep{liu2024rlgpt}) operate in incompatible paradigms that preclude controlled comparison (please refer to Appendix~\ref{app:taxonomy} for discussion).


\subsection{Main Results} 
\begin{figure}[t]
    \centering
    \includegraphics[width=0.85\linewidth]{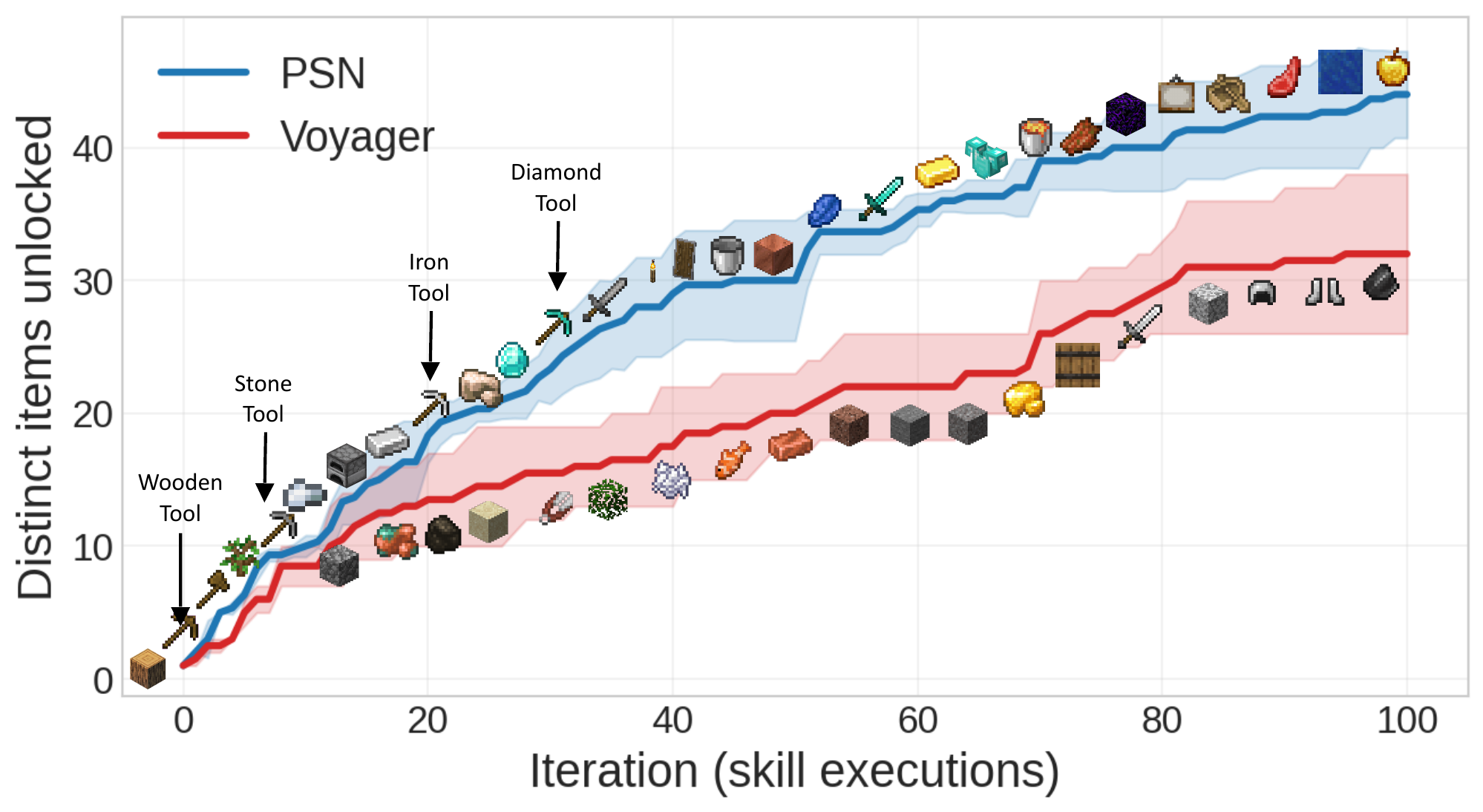}
    \caption{Tech tree mastery on Minecraft.}
    \label{fig:minecraft_techtree}
\end{figure}

\paragraph{Minecraft Tech Tree Mastery.} 
Figure~\ref{fig:minecraft_techtree} and Table~\ref{tab:minecraft_main} compare agents in terms of technology tree progression, measured by the number of iterations. Progressing along the tech tree requires solving increasingly long-horizon and compositional tasks, where later-stage tools depend on reliable execution and reuse of earlier skills.
PSN exhibits substantially faster and more stable progression than all baselines. ReAct and Reflexion fail to unlock any tool-level milestones. AutoGPT completes early-stage objectives but struggles to sustain progress beyond iron-level tools, exhibiting high variance. Voyager achieves consistent progress through iron tools, but slows significantly at the diamond stage.
In contrast, PSN continues to unlock higher-tier items with fewer attempts and lower variance, indicating that persistent programmatic skills, trace-based credit assignment, and structural refactoring enable sustained long-horizon competence. For obsidian acquisition, PSN executes a multi-step procedure (i.e., bucket crafting, water-lava interaction, and diamond-pickaxe mining) encapsulated as a single composed skill that extensively reuses previously learned subskills, illustrating PSN's ability to compress long-horizon behaviors into reusable programmatic abstractions.
A detailed comparison positioning PSN, ADAM, and ODYSSEY along an autonomy--engineering spectrum is provided in Appendix~\ref{app:odyssey_comparison}. Crucially, these results are consistent across model families: under Qwen3-Coder-Next (3B active parameters), PSN reaches diamond tool in all 6 runs while Voyager$^\dagger$ succeeds in only 1/6 (Table~\ref{tab:minecraft_main}), confirming that the gains arise from PSN's architectural scaffolding rather than the specific LLM backend. By contrast, ADAM$^\dagger$'s causal inference degrades sharply under Qwen3, failing beyond stone tools, as its memoryless causal discovery mechanism is sensitive to LLM reasoning quality. A direct ablation of the inline knowledge scaffolding (Appendix~\ref{app:scaffolding_ablation}) further verifies this attribution at the prompt level: on GPT-5-mini, the inline knowledge corrigendum (the \texttt{tool\_tier\_rules} item) contributes only $+2$pp to diagnostic correctness on Phase~1 reflection cases (98\% without it vs near-100\% with it); on Qwen3, the same weak-LLM corrigendum contributes $+33$pp on synthetic tier-mismatch failures and is documented as such in the codebase.

\begin{figure}[t]
    \centering
    \includegraphics[width=0.85\linewidth]{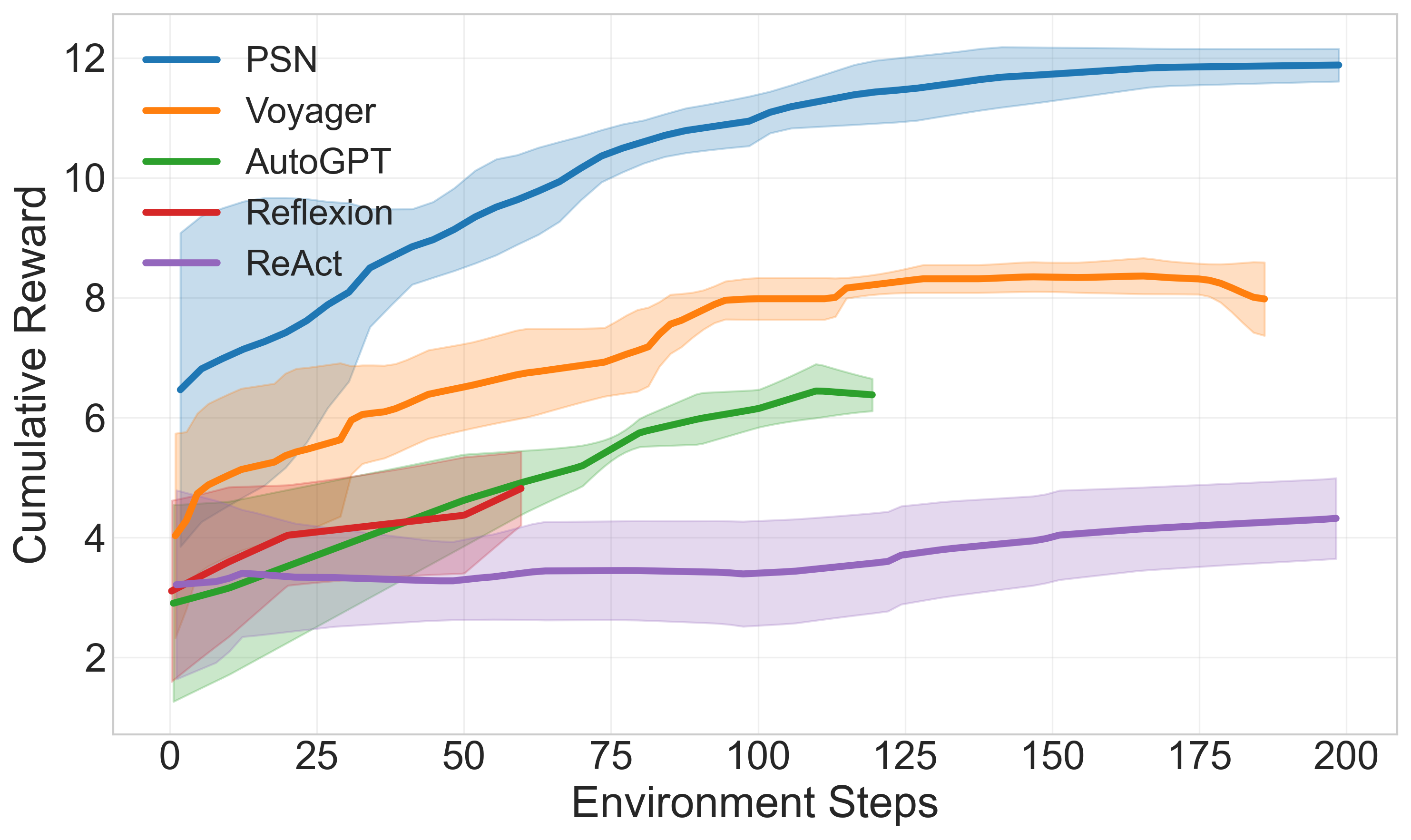}
    \caption{Cumulative Reward on Crafter. Shorter curves indicate earlier \textbf{agent death} due to Crafter's survival mechanics (hostile mobs, hunger, hazards).}
    \label{fig:crafter_reward}
\end{figure}

\paragraph{Crafter.}
Figure~\ref{fig:crafter_reward} reports cumulative episode reward on Crafter, which reflects the agent's ability to survive, gather resources, and make progress under dense feedback. Shorter curves indicate earlier agent death due to Crafter's survival mechanics (hostile mobs, hunger, hazards). Unlike Minecraft benchmarks emphasizing sparse milestone completion, Crafter requires sustained stability where early mistakes can compound. PSN consistently achieves higher reward. Voyager achieves stabler returns than planning-only baselines, but remains limited by its flat skill library. By contrast, PSN maintains steadily increasing reward throughout training, showing broader generalization to dense continual learning.

\begin{figure}[t]
    \centering
    \includegraphics[width=0.85\linewidth]{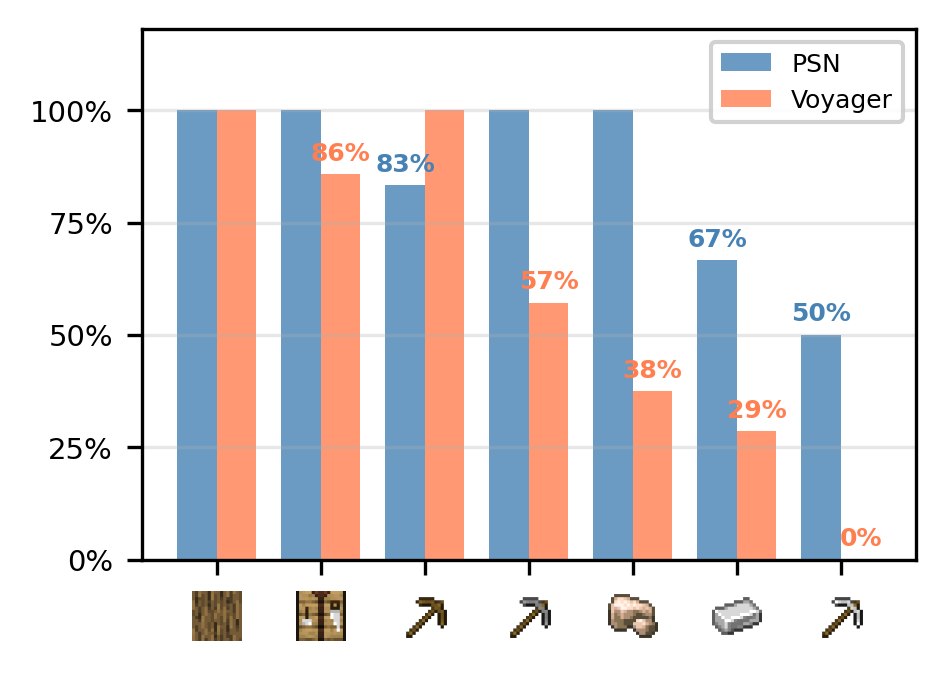}
    \caption{Skill Retention Rate under continual learning setting. PSN consistently preserves previously mastered skills, while Voyager exhibits severe catastrophic forgetting as training progresses.}
    \label{fig:skill_retention}
\end{figure}


\subsection{Generalization}
\paragraph{Continual Learning over Task Streams (Temporal Generalization).}
Since the continual skill acquisition efficiency of PSN can be observed in Figure~\ref{fig:minecraft_techtree},
we evaluate PSN's ability to acquire increasingly complex skills from a sequential task stream while avoiding catastrophic forgetting. Tasks are presented in a fixed curriculum following the technology tree (see Appendix~\ref{app:task_sequences} for full sequences). Each task is trained until its success rate exceeds a predefined threshold (marked as mastered), or until a maximum number of attempts is reached. To measure forgetting, we introduce the \emph{Skill Retention Rate (SRR)}: once a task is mastered, it is periodically re-evaluated after each subsequent task is mastered, and SRR is defined as the cumulative success rate across all such re-evaluations. As shown in Figure~\ref{fig:skill_retention}, PSN consistently preserves earlier skills as training progresses, whereas Voyager exhibits severe backward interference, with retention rapidly degrading as new skills are learned. These results demonstrate that structured credit assignment and maturity-aware stabilization are critical for continual skill acquisition.


\paragraph{Compositional Generalization via Network-Aware Skill Reuse.}
We hypothesize that PSN solves unseen compositional tasks by reusing and recombining existing skills rather than synthesizing new ones. To test this, we introduce a controlled baseline, PSN (Create New Skills), which bypasses backward chaining and always synthesizes a new skill for each task. Figure~\ref{fig:voyager_growth} compares skill repertoire sizes as training progresses. Early in training, both variants grow similarly as foundational skills are acquired. However, the gap widens over time: PSN's repertoire plateaus while PSN (Create New Skills) continues to accumulate skills. This indicates that PSN increasingly grounds new tasks in its existing skill network via backward chaining, achieving compositional generalization through reuse rather than proliferation. Notably, PSN's repertoire even decreases in later iterations, suggesting that the refactoring mechanism actively merges redundant helper functions over time.


\subsection{Ablation Study}
\paragraph{End-to-End Optimizer.}
We ablate the symbolic optimizer to disentangle the effect of optimization from that of skill representation. As shown in Table~\ref{tab:minecraft_main}, PSN without the optimizer achieves performance comparable to Voyager on early- and mid-stage tools (wooden, stone, and iron). However, this variant fails to reliably progress to later-stage objectives such as diamond tools and obsidian, mirroring Voyager's degradation under increasing task depth. In contrast, the full PSN consistently unlocks higher-tier items with substantially fewer iterations. This gap indicates that the optimizer is not required to make skills functional, but is critical for repairing brittle behaviors and enabling stable scaling to long-horizon, deeply compositional tasks.

\begin{figure}[t]
    \centering
    \includegraphics[width=0.85\linewidth]{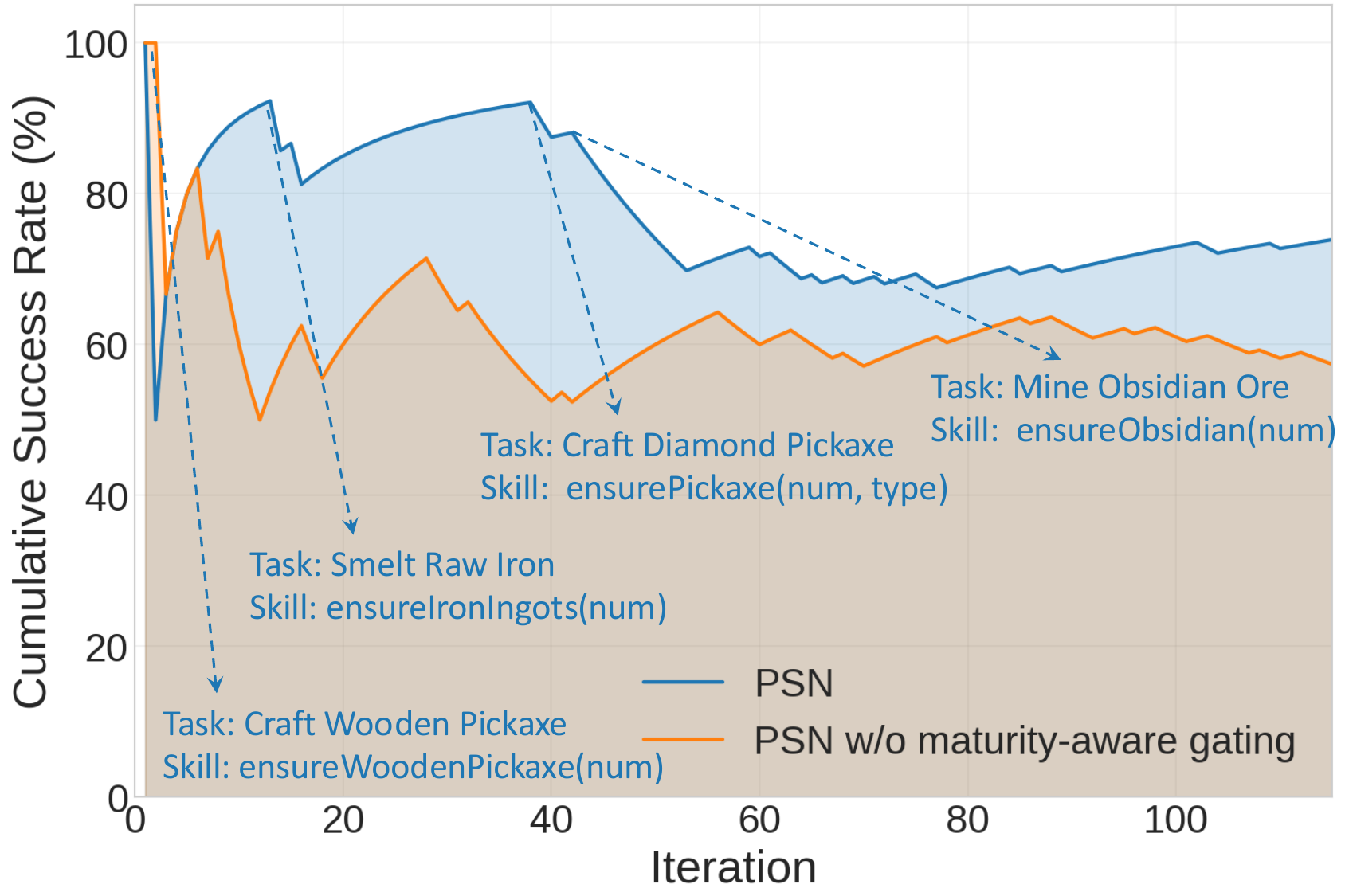}
    \caption{The cumulative task success rate of PSN w/ and w/o maturity gating.}
    \label{fig:voyager_stabaliation}
\end{figure}

\paragraph{Maturity-aware update gating gradually stabilizes learned skills.}
Figure~\ref{fig:voyager_stabaliation} compares cumulative success rates for PSN with and without maturity-aware update gating. Without stabilization, converged skills are repeatedly modified by downstream failures, leading to oscillatory behavior. By contrast, maturity-aware gating progressively reduces the update frequency of reliable skills while allowing immature skills to remain plastic. As a result, PSN with stabilization achieves higher cumulative success rates and stabler learning dynamics.

\begin{figure}[t]
    \centering
    \includegraphics[width=0.85\linewidth]{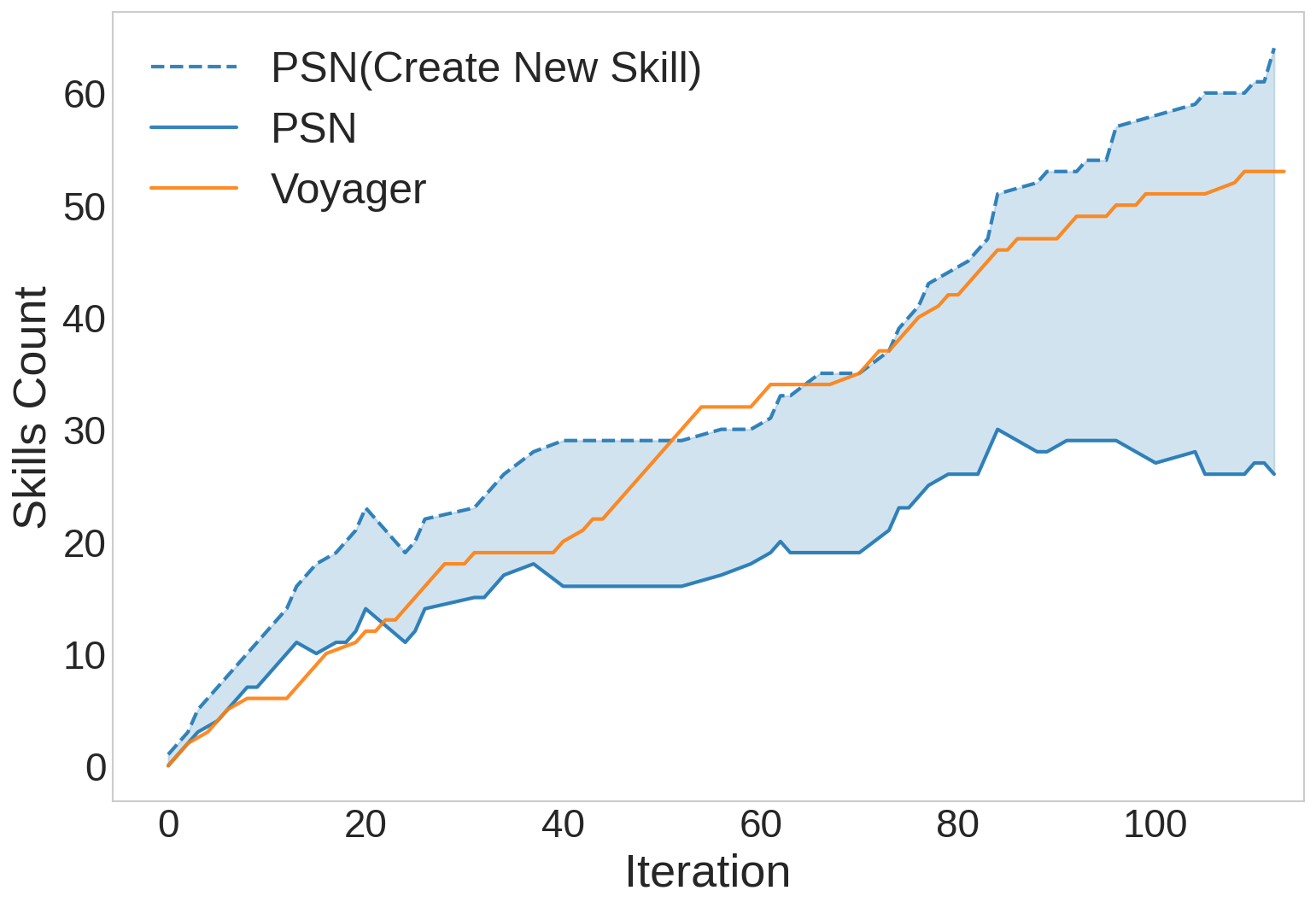}
    \caption{Growth of the skill library over training. In PSN (Create New Skills), the agent always synthesizes a new skill for each task. Compared to baselines, PSN reuses and optimizes existing skills, maintaining a compact skill repertoire.}
    \label{fig:voyager_growth}
\end{figure}

\paragraph{Refactor Regulates the Network Growth.}
Figure~\ref{fig:voyager_growth} shows how the size of the skill library evolves as learning progresses. In PSN (Create New Skills), the agent always synthesizes a new skill for each task. Without structural refactoring, Voyager's skill library grows rapidly, accumulating redundant or overly specialized skills. This uncontrolled growth increases planning complexity and degrades efficiency. In contrast, PSN maintains a significantly more compact skill network by identifying canonical redundancy patterns and applying semantics-preserving rewrites. As a result, the effective growth rate is substantially reduced even as task complexity increases.

\paragraph{Offline Refactor vs. Online Refactor.}
To test whether structural compression alone is sufficient, we apply an \emph{offline refactor} to Voyager's learned skill library using a strong LLM (Claude Opus~4.5), which refactored its 58 existing skills into 7 generic skills, 20 lightweight wrappers, and 38 unchanged skills (65 total), denoted as Voyager-R. While this offline refactoring significantly reduces redundancy (in terms of repeating code blocks), it does not yield the same behavioral robustness. When evaluated on a fixed sequence of compositional tasks (Appendix~\ref{app:task_sequences}),
Voyager-R achieves a success rate of 0.6875, compared to 0.8462 for PSN with online refactoring. This gap indicates that refactoring is most effective when performed \emph{online} and tightly coupled with execution feedback, rather than applied once to a static skill library.

\subsection{REFLECT under LLM noise: a three-layer view}
\label{sec:reflect_noise}
\begin{figure}[t]
\centering
\includegraphics[width=\linewidth]{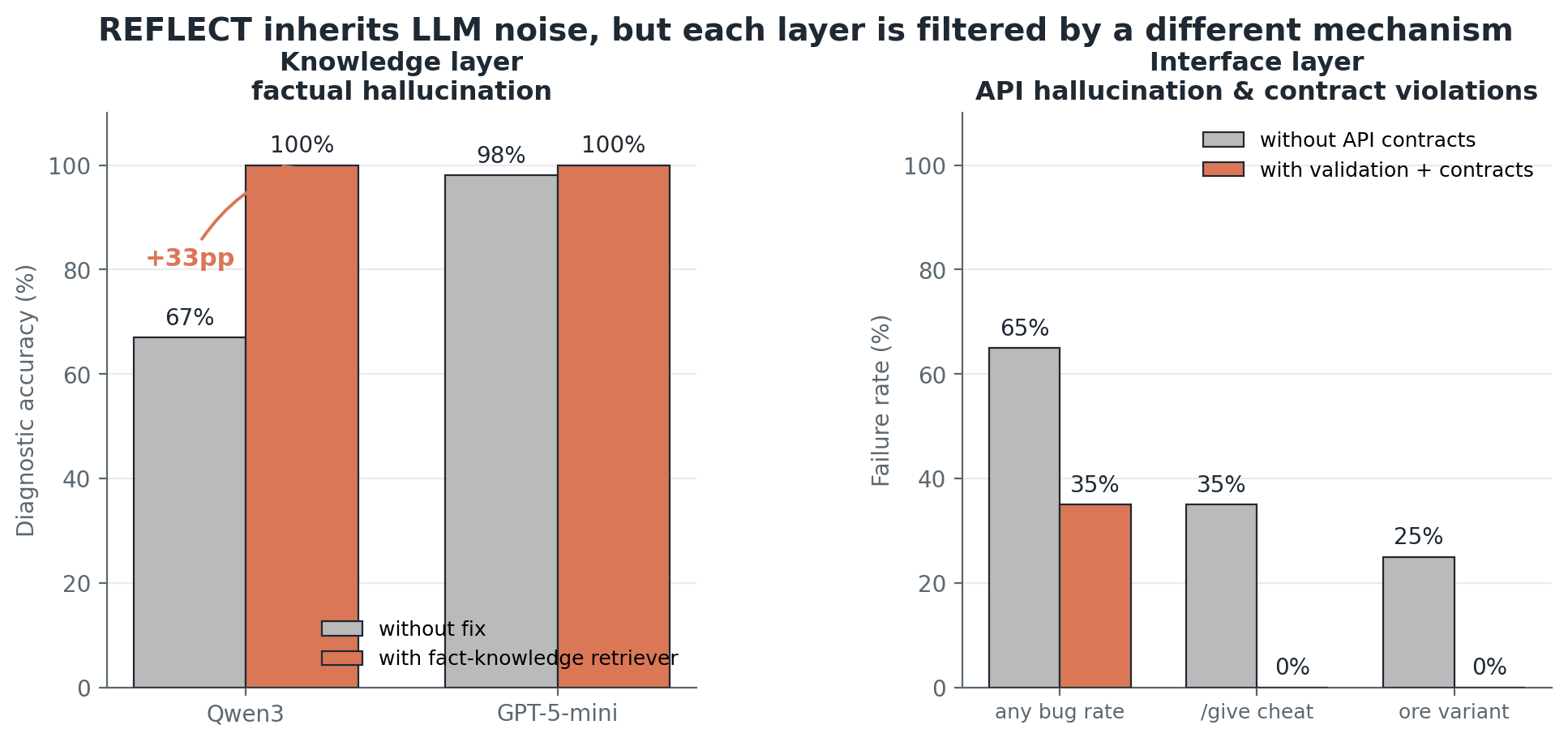}
\caption{REFLECT inherits LLM noise, but each layer is filtered by a different mechanism. \textbf{Left (knowledge layer):} factual hallucination is mitigated by a model-specific \texttt{tool\_tier\_rules} corrigendum, worth $+33$pp on Qwen3 but only $+2$pp on GPT-5-mini. \textbf{Right (interface layer):} fabricated or misused APIs are caught before execution by a static reference checker plus API contracts, lowering the bug rate across both LLMs. The reasoning layer is handled by env-validated iteration (Table~\ref{tab:srk}).}
\label{fig:noise_layers}
\end{figure}

\opreflect is instantiated by a prompted LLM, so its repair signal is noisy by construction. The question is not how to make it noise-free, which is impossible, but how the surrounding architecture keeps learning stable despite it. We decompose REFLECT noise into three layers and locate the source of tolerance in each (Figure~\ref{fig:noise_layers}).

\paragraph{Knowledge layer (missing/incorrect facts).} The model misremembers a domain fact needed for diagnosis (e.g.\ Qwen3 asserting ``raw\_iron does not exist in vanilla Minecraft''). This is the only layer whose mitigation is model-specific: on synthetic tier-mismatch diagnoses, one injected \texttt{tool\_tier\_rules} corrigendum contributes $+33$pp on Qwen3 but only $+2$pp on GPT-5-mini (Appendix~\ref{app:scaffolding_ablation}). Knowledge-layer noise therefore all but vanishes on a stronger model: the corrigendum is a pretraining patch, not an architectural mechanism.

\paragraph{Reasoning layer (mis-attributed causes).} With the right facts available, the model can still build a plausible but wrong causal story (e.g.\ blaming lava for an obsidian non-drop that is actually a tool-tier mismatch). No per-call fix exists; tolerance is architectural and operates at the loop level. Because every diagnosis is applied and re-executed ($s_{i+1}=s_i+\delta s_i$), a wrong fix fails its effect-check and is re-diagnosed against fresh evidence. This env-validated iteration turns single-shot diagnostic noise into a convergent process: first-attempt task success of 67\% rises to 100\% within three attempts (Table~\ref{tab:srk}).

\paragraph{Interface layer (fabricated/misused APIs).} The model emits code that calls non-existent methods or misuses real ones. PSN gates this before execution with a static reference checker (Babel parse and symbol resolution) plus explicit API-contract documentation, so fabricated calls are rejected rather than cascading into the trace. On a production-prompt replay ($n{=}20$), this gate lowers the interface-layer bug rate from 65\% to 35\%, an effect that holds across LLMs.

Two of the three defenses, the env-validated loop and the static interface gate, are architectural and work across LLMs; only the knowledge corrigendum is model-specific, and it matters mainly for the weaker model.

\subsection{Cross-model robustness and compositional reuse}
\label{sec:cross_llm}
PSN reaches the diamond tool on both a strong closed model (GPT-5-mini) and a weaker open mixture-of-experts model (Qwen3-Coder-Next, 3B active / 80B total), in $35\pm16$ and $49\pm18$ iterations over six runs each, with no architectural changes (Table~\ref{tab:minecraft_main}). The two models differ sharply in capability, yet three independent measurements all shift with model strength while the same architecture absorbs the difference.

\begin{figure}[t]
\centering
\begin{subfigure}[t]{0.49\linewidth}
\centering
\includegraphics[width=\linewidth]{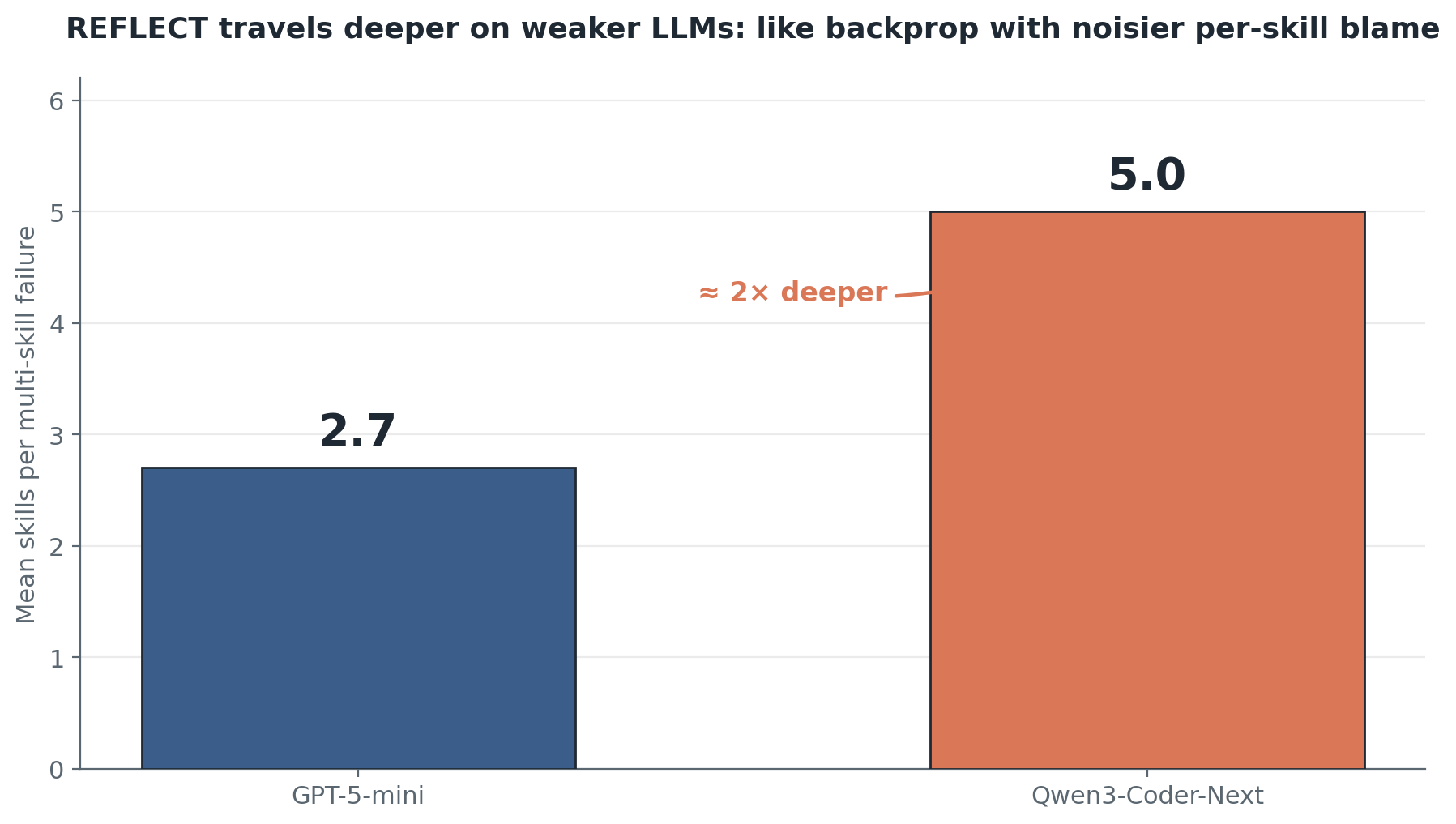}
\caption{REFLECT propagation depth scales with model strength: the weaker model propagates failure feedback through $5.0$ skills per multi-skill repair on average, versus $2.7$ for the stronger one, consistent with the chain-rule view of credit assignment (\S\ref{sec:optimization}).}
\label{fig:reflect_depth}
\end{subfigure}\hfill
\begin{subfigure}[t]{0.49\linewidth}
\centering
\includegraphics[width=\linewidth]{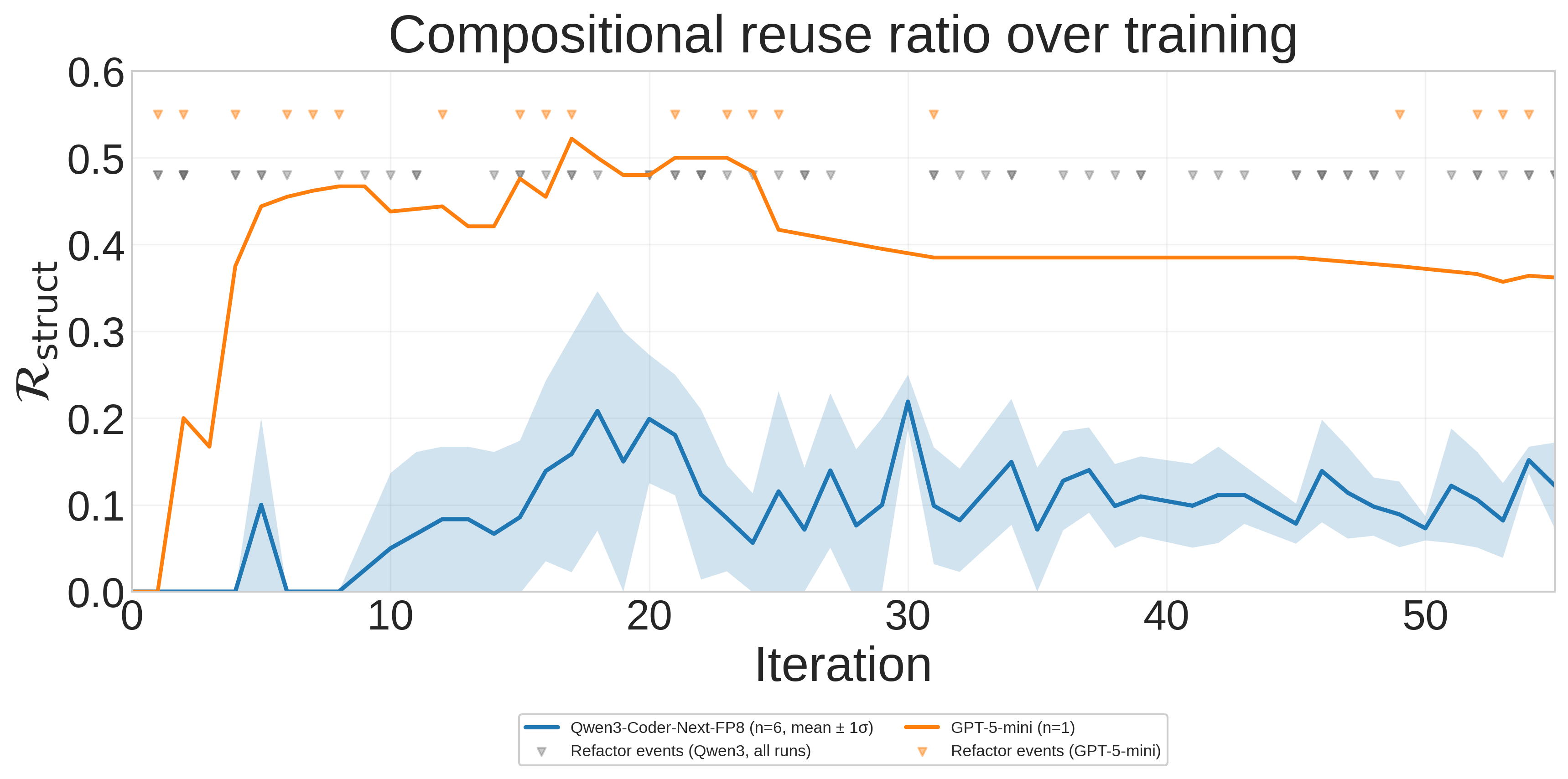}
\caption{Compositional reuse ratio $\mathcal{R}_{\text{struct}}$ over training. The stronger model (GPT-5-mini, single hero run) climbs toward $\approx0.4$; the weaker model (Qwen3-Coder-Next, mean$\pm1\sigma$ over six runs) plateaus markedly lower.}
\label{fig:rstruct_traj}
\end{subfigure}
\caption{Two of the three model-dependent signals in \S\ref{sec:cross_llm}: deeper credit-assignment chains (left) and lower compositional reuse (right) on the weaker model.}
\label{fig:cross_llm_signals}
\end{figure}

\paragraph{Three signals scale with the model.} (i)~\emph{REFLECT depth}: the weaker model needs deeper credit assignment, propagating failure feedback through $5.0$ skills per multi-skill repair on average, versus $2.7$ for the stronger model (Figure~\ref{fig:reflect_depth}). (ii)~\emph{Knowledge dependence}: the \texttt{tool\_tier\_rules} corrigendum contributes $+33$pp to the weaker model's diagnostic correctness but only $+2$pp to the stronger one's (\S\ref{sec:reflect_noise}). (iii)~\emph{Compositional reuse}: the reuse ratio $\mathcal{R}_{\text{struct}}$, the fraction of skills with fan-in $>1$, asymptotes near $0.4$ for the stronger model but only $0.15\pm0.07$ for the weaker one (Figure~\ref{fig:rstruct_traj}). A weaker model thus emits a noisier repair signal, leans harder on explicit knowledge, and grows a flatter, less-reused network.

\paragraph{The architecture absorbs the noise.} None of these differences breaks learning: both models complete the tech tree under the same loop, gating, and refactor operators. The weaker model simply works harder, diagnosing deeper and reusing less, while the surrounding training stack converts its noisier signal into the same outcome. Reliability comes from the architecture rather than from the model being strong.

\begin{figure}[t]
\centering
\includegraphics[width=0.8\linewidth]{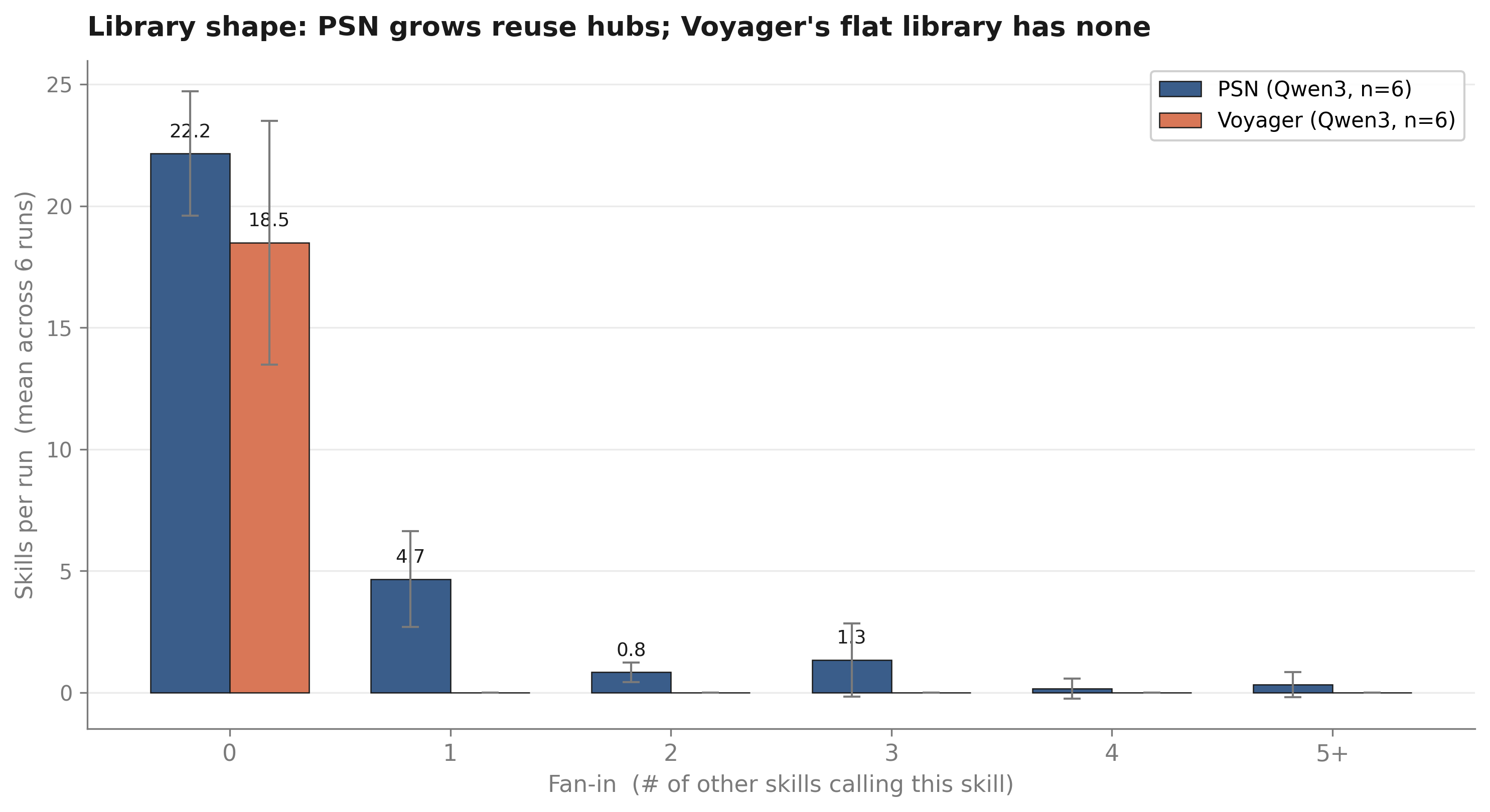}
\caption{Library shape under the weaker model (Qwen3-Coder-Next), averaged over six runs each. PSN grows reuse hubs (skills with fan-in $\geq1$), whereas Voyager's flat retrieval library nearly keeps every skill at fan-in $0$.}
\label{fig:fan_in}
\end{figure}

\paragraph{Reuse topology.} Compositional reuse is also visible in the shape of the learned graph. PSN develops a hub-and-spoke structure in which a few utility skills (e.g.\ \texttt{setupCraftingTable} and \texttt{ensureResource}) are invoked by many parents, whereas a flat retrieval library such as Voyager's nearly keeps every skill at fan-in $0$ by construction (Figure~\ref{fig:fan_in}).

\section{Related Work}
\label{sec:related_work}

\noindent\textbf{Skill Learning and Hierarchical RL.}
Hierarchical RL studies temporal abstraction via options \citep{sutton1999options,barto2003recent,bacon2017optioncritic,eysenbach2019diayn} and modular routing \citep{andreas2016nmn,xu2018neural,zhang2018composable,shazeer2017moe,riquelme2021routing}. LLM-guided approaches segment trajectories into reusable skills via variational inference \citep{fu2024last}. Unlike these work, PSN represents skills as executable programs with explicit control flow and pre/postconditions.

\noindent\textbf{LLM-based Agents and Program Synthesis.}
LLM agents maintain code memories or skill repositories \citep{yao2023react,schick2023toolformer,ahn2022saycan,wang2024voyager,prabhu2025walt}. CodeAct \citep{wang2024codeact} uses executable code as a unified action space; ReGAL \citep{stengel-eskin2024regal} learns function libraries via refactoring, capturing environment dynamics; MINDcraft \citep{white2025collaborating} studies multi-agent task solving; ASI \citep{wang2025asi} induces programmatic skills on-the-fly for web agents; AgentCoder \citep{huang2023agentcoder} uses multi-agent code generation; DiVE \citep{sun-etal-2024-enhancing-agent} builds natural language knowledge libraries. \citet{wang2025bytesized32refactored} show refactoring facilitates coding agents. Self-improving agents learn via RL-based skill accumulation \citep{wang2025reinforcement}, reasoning memory \citep{ouyang2025reasoningbank}, or skill disclosure \citep{anthropic2025agentskills}. ADAM~\citep{zhang2024adam} learns causal input-output mappings for 47 predefined skills but does not generate or optimize code; ODYSSEY~\citep{liu2024odyssey} relies on 183 hand-authored skills with LLM-based retrieval. Differently, PSN autonomously generates, composes and optimizes skill code within a compositional network, without hand-authored task-specific skills.

\noindent\textbf{Neuro-Symbolic Learning and Architecture Optimization.}
Neuro-symbolic systems integrate symbolic structures with differentiable computation \citep{garcez2019neural,baydin2018automatic,badreddine2022logic,manhaeve2018deepproblog}. OneLife \citep{khan2025one} models dynamics via programmatic laws with precondition-effect structures, analogous to PSN's skill representation. Symbolic-MoE \citep{chen2025symbolic} routes through skill-based experts; EFA \citep{khan2025executable} infers executable abstractions for math. Neural architecture search prunes and restructures networks \citep{zoph2017nas,han2016deepcompression,tan2019efficientnet}, with techniques like learning rate scheduling enabling stability-plasticity tradeoffs \citep{howard2018ulmfit,yosinski2014transfer,rusu2016progressive}. PSN draws on both traditions: it embeds learning dynamics inside symbolic programs rather than in differentiable models, while performing architecture-search-like refactoring under rollback validation.

\section{Conclusion}
We introduced PSN, a framework for continual skill acquisition where executable symbolic programs form a compositional network that evolves through experience. PSN's three mechanisms (i.e., trace-based credit assignment, maturity-aware update gating, and canonical structural refactoring) induce learning dynamics with structural parallels to neural network training. Experiments on Minecraft and Crafter demonstrated faster skill acquisition, reduced forgetting, and superior compositional generalization, suggesting that principles from neural network optimization can inform the design of symbolic learning systems.

\section*{Ethics Statement}
This work studies autonomous skill acquisition in simulated game environments (Minecraft and Crafter). The agents operate in sandboxed simulators with no real-world deployment, posing no direct safety risks. Our framework uses prompted LLMs as code generation backends; we do not fine-tune models or train on private data. All experiments use publicly available game environments and APIs. We acknowledge that autonomous code generation systems could, in principle, produce unintended behaviors; PSN mitigates this through rollback validation and execution sandboxing.

\section*{Reproducibility Statement}
To ensure reproducibility, we provide: (1)~complete algorithmic specifications of all PSN operators in the main text and appendix, including the two-phase optimization algorithm (Appendix~\ref{app:two_phase_optimizer}); (2)~example prompt templates for all LLM-instantiated operators (Appendix~\ref{app:prompt_templates}); (3)~exact model identifiers (\texttt{gpt-5-mini-2025-08-07} and \texttt{Qwen3-Coder-Next} with 3B active / 80B total parameters); (4)~all hyperparameters ($\gamma{=}5.0$, $\epsilon{=}0.1$, maturity pivot $0.6$, rollback threshold 20\%, momentum window 5); (5)~task sequences used for evaluation (Appendix~\ref{app:task_sequences}); and (6)~detailed code diffs for representative optimization cases (Appendix~\ref{app:optimization-diffs}). All experiments are conducted on Minecraft 1.19.4 via MineDojo/Mineflayer and Crafter.

\bibliography{colm2026_conference}
\bibliographystyle{colm2026_conference}

\clearpage
\appendix
\label{sec:appendix}

\section*{Limitations}
\label{limitations}
Our current implementation of PSN operates under constrained computational resources, resulting in an effectively batch-size-one online learning regime. This significantly limits the degree of parallelism in both skill execution and reflection-driven optimization, and prevents us from fully exploring large-scale network-level learning dynamics.

Moreover, the current reflection and refactoring process lacks a formal projection guarantee in the symbolic program space. While empirical improvements are consistently observed, the theoretical properties of symbolic projection, convergence, and optimality remain to be established.

Nevertheless, we believe these limitations are not fundamental to the PSN paradigm. With the continued scaling of large language models, increased computational budgets, and more efficient parallel execution infrastructures, future iterations of PSN are expected to support large-batch learning, stronger theoretical guarantees, and substantially improved optimization efficiency.

\section{Two-Phase Optimization Algorithm of Skill Optimizer}
\label{app:two_phase_optimizer}
This section provides a formal algorithmic specification of the two-phase skill optimization process described in the main paper. Algorithm~\ref{alg:two_phase_skill_opt} summarizes the complete procedure. A key distinction in our framework is between a skill’s \emph{feedback} and its \emph{gradients}:
feedback indicates \emph{what went wrong}, while gradients encode \emph{how the skill should be modified}.

\subsection{Feedback vs. Gradients}
For a skill $s$, we denote by $f_s$ the feedback signal assigned to $s$ after task execution.
This feedback may arise from task failure, unmet subgoals, or trace-level diagnostics. Crucially, $f_s$ does not directly specify how to modify $s$.

Instead, PSN performs a symbolic analysis step that converts feedback into gradients.
We denote this process as:
\[
\textsc{Reflect}(s, f_s, \mathrm{Subskill}(s))
\;\rightarrow\;
\bigl(g_s,\; \{f_{s'}\}_{s' \in \mathrm{Subskill}(s)}\bigr),
\]
where $g_s$ (also written as $\tilde{\nabla}_s$) is a gradient-like modification proposal for $s$,
and $f_{s'}$ are newly generated feedback signals for each sub-skill invoked by $s$.

This operation implements a symbolic form of differentiation over the skill invocation structure.

\subsection{Phase I: Top-down Feedback Backpropagation}

Phase I performs \emph{top-down feedback backpropagation} over the skill network. Starting from a skill that fails to complete a task, PSN recursively applies \textsc{Reflect} following the invocation relations induced by the execution trace.

At each skill $s$, symbolic differentiation decomposes $f_s$ into:
\begin{itemize}[leftmargin=1.2em]
    \item a local gradient proposal $g_s$ describing how $s$ itself should be modified, and
    \item feedback signals $\{f_{s'}\}$ assigned to sub-skills $s' \in \mathrm{Subskill}(s)$.
\end{itemize}

This process continues until no further sub-skills require feedback propagation. The result of Phase I is a \emph{pending optimization subgraph} consisting of:
\[
\mathcal{G}_{\text{opt}} = \{(s, g_s)\},
\]
i.e., a connected subgraph of skills paired with their gradient proposals. No skill code is modified during this phase.

\subsection{Phase II: Bottom-up Gradient Application}

Phase II applies gradients in a \emph{bottom-up} manner over $\mathcal{G}_{\text{opt}}$.
Skills are updated in an order that respects dependency relations, starting from leaf skills
and proceeding toward higher-level skills.

For a skill $s$ with gradient proposal $g_s$, the update is performed via:
\[
\textsc{ApplyGradients}\bigl(s,\; g_s,\; \mathcal{C}_s\bigr).
\]

Here, $\mathcal{C}_s$ is a \emph{context object} that aggregates optimization reports
returned by sub-skills that have already been updated.
Let
\[
\mathcal{S}_s \;:=\; \mathrm{Subskill}(s)
\]
denote the set of sub-skills invoked by $s$.
The context $\mathcal{C}_s$ is constructed as:
\[
\mathcal{C}_s \;:=\;
\textsc{Consider}\bigl(
\textsc{OptimizeReport}(\mathcal{S}_s)
\bigr),
\]
which summarizes feedback signals derived from the updated sub-skills.

Updates are realized through program-level rewrite, patch, or diff operations on the skill code.
After updating $s$, the optimizer generates an \emph{optimization report} summarizing the changes
and their effects. This report is propagated upward and used to inform subsequent updates of parent skills, allowing higher-level skills to adapt consistently to changes in their dependencies.

\subsection{Algorithmic Interpretation}
The complete optimization step thus consists of two strictly separated phases:
\begin{itemize}[leftmargin=1.5em]
    \item \textbf{Phase I:} Top-down symbolic differentiation to propagate feedback $\{f_s\}$.
    \item \textbf{Phase II:} Bottom-up application of gradient proposals $\{g_s\}$.
\end{itemize}

This design explicitly decouples \emph{credit assignment} from \emph{code modification}. While Phase I follows a chain-rule-like decomposition of feedback signals, Phase II ensures that updates are applied in a dependency-consistent order, preventing interference between skills during optimization.

\subsection{Discussion}
By separating feedback propagation from gradient application, PSN generalizes the backward--forward separation of neural backpropagation to symbolic, programmatic skill networks. We find this two-phase structure essential for stable optimization in deeply compositional and long-horizon tasks.

\begin{algorithm*}[t]
\caption{Two-Phase Skill Optimization in PSN
(\emph{Phase I}: top-down feedback backpropagation;
\emph{Phase II}: bottom-up gradient application)}
\label{alg:two_phase_skill_opt}

\KwIn{Root skill $s_{\mathrm{root}}$, task feedback $f_{s_{\mathrm{root}}}$, execution trace $\mathcal{T}$}
\KwOut{Updated skills and optimization reports}

\BlankLine
\textbf{Definitions.}
$\mathrm{Subskill}(s;\mathcal{T})$: sub-skills invoked by $s$ in $\mathcal{T}$\;
$\textsc{Reflect}(s,f_s,\mathrm{Subskill}) \rightarrow (g_s,\{f_{s'}\})$\;
$\textsc{ApplyGradients}(s,g_s,\mathcal{C}) \rightarrow (s^{+},r_s)$\;

\BlankLine
\textbf{Phase I: Top-down feedback backpropagation (symbolic differentiation).}

Initialize maps $\mathcal{G}\gets\emptyset$ (gradients), $\mathcal{F}\gets\emptyset$ (feedback)\;
Initialize queue $Q\gets[(s_{\mathrm{root}},f_{s_{\mathrm{root}}})]$\;

\While{$Q\neq\emptyset$}{
    Pop $(s,f_s)$ from $Q$\;
    $\mathcal{F}[s]\gets f_s$\;
    $\mathcal{S}\gets\mathrm{Subskill}(s;\mathcal{T})$\;
    $(g_s,\{f_{s'}\}_{s'\in\mathcal{S}})\gets\textsc{Reflect}(s,f_s,\mathcal{S})$\;
    $\mathcal{G}[s]\gets g_s$\;
    \ForEach{$s'\in\mathcal{S}$}{
        \If{$f_{s'}\neq\varnothing$}{
            Push $(s',f_{s'})$ into $Q$\;
        }
    }
}
Let $\mathcal{H}$ be the induced pending optimization subgraph over $\mathrm{Dom}(\mathcal{G})$\;

\BlankLine
\textbf{Phase II: Bottom-up gradients application (dependency-respecting updates).}

Compute bottom-up order $\pi\gets\textsc{PostOrder}(\mathcal{H})$\;
Initialize report map $\mathcal{R}\gets\emptyset$\;

\ForEach{$s$ in $\pi$}{
    $\mathcal{C}\gets\textsc{Consider}(\{\textsc{OptimizeFeedback}(s')\mid s'\in\mathrm{Subskill}(s)\cap\mathrm{Dom}(\mathcal{R})\})$\;
    $(s^{+},r_s)\gets\textsc{ApplyGradients}(s,\mathcal{G}[s],\mathcal{C})$\;
    Replace $s\leftarrow s^{+}$ in the skill net\;
    $\mathcal{R}[s]\gets r_s$\;
}

\Return{$\{s^{+}\}$ and $\mathcal{R}$}\;

\end{algorithm*}

\begin{table*}[t]
\centering
\small
\setlength{\tabcolsep}{5pt}
\renewcommand{\arraystretch}{1.12}
\resizebox{\textwidth}{!}{%
\begin{tabular}{p{0.05\textwidth}p{0.18\textwidth}p{0.57\textwidth}p{0.12\textwidth}}
\toprule
\textbf{Case} & \textbf{Pattern} & \textbf{Example and rewrite} & \textbf{Illustration} \\
\midrule
(A) & Parametric coverage &
\textbf{Example:} \texttt{mineLogs(type,num)} generalizes \texttt{mineOakLogs(num)}.
\textbf{Rewrite:} \texttt{mineOakLogs(num) := mineLogs(OAK,num)}. &
Figure~\ref{fig:refactor-parametric} \\

(B) & Behavioral / subgraph coverage &
\textbf{Example:} \texttt{craftCraftingTable} inlines routines that exist as skills.
\textbf{Rewrite:} replace duplicated blocks by calls to \texttt{mineLogs} and \texttt{craftPlanks}. &
Figure~\ref{fig:refactor-subgraph}\\

(C) & Sibling specializations &
\textbf{Example:} \texttt{mineOakLogs(num)} and \texttt{mineBirchLogs(num)} indicate a missing abstraction.
\textbf{Rewrite:} synthesize \texttt{mineLogs(type,num)} and rewrite both as wrappers. &
Figure~\ref{fig:refactor-sibling}\\

(D) & Extract common subskill &
\textbf{Example:} both \texttt{craftSticks} and \texttt{craftTable} require \texttt{ensurePlanks(k)}.
\textbf{Rewrite:} extract \texttt{ensurePlanks(k)} as a new skill and replace both occurrences by a call. &
Figure~\ref{fig:refactor-common}\\

(E) & Duplication &
\textbf{Example:} two skills are near-identical up to naming/surface variations.
\textbf{Rewrite:} keep higher-$V(s)$ canonical skill; redirect incoming links; demote the other to an alias. &
Figure~\ref{fig:refactor-duplication} \\
\bottomrule
\end{tabular}%
}
\caption{Index of canonical refactor cases supported by PSN. Each case corresponds to a distinct structural relationship and rewrite rule, with detailed illustrations provided in Appendix~\ref{app:refactor-casebook}.}
\label{tab:refactor-casebook}
\end{table*}

\section{Refactor Casebook}
\label{app:refactor-casebook}
This appendix presents a visual casebook of the canonical refactor patterns supported by the Programmatic Skill Network (PSN). Each case corresponds to a distinct structural relationship between skills and induces a deterministic graph rewrite. All cases referenced in Section~\ref{subsec:refactor} are illustrated in the Table~\ref{tab:refactor-casebook} and below.

These refactor cases are exhaustive with respect to the structural patterns observed in our experiments.

\subsection{Case A: Parametric Coverage}
\label{app:refactor-parametric}

\begin{figure*}[t]
\centering
\includegraphics[width=0.88\textwidth]{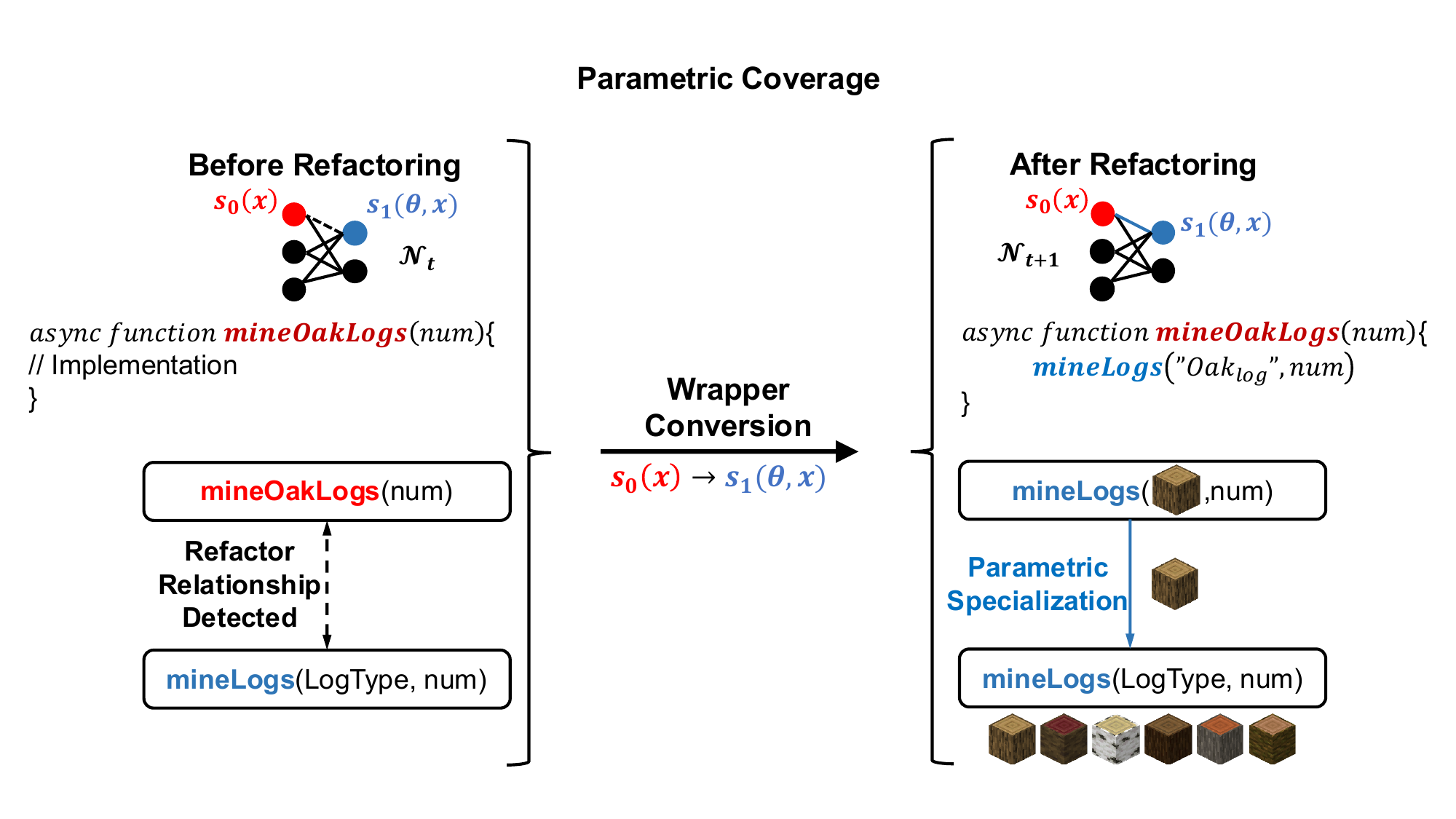}
\caption{Parametric coverage.
A specialized skill is rewritten as a wrapper around a more general, parameterized skill.}
\label{fig:refactor-parametric}
\end{figure*}

\paragraph{Pattern.}
One skill is a strict specialization of another skill that admits a parameterized generalization.

\paragraph{Rewrite.}
The specialized skill is replaced by a thin wrapper that calls the generalized skill with fixed parameter values.

\subsection{Case B: Behavioral / Subgraph Coverage}
\label{app:refactor-subgraph}

\begin{figure*}[t]
\centering
\includegraphics[width=0.88\textwidth]{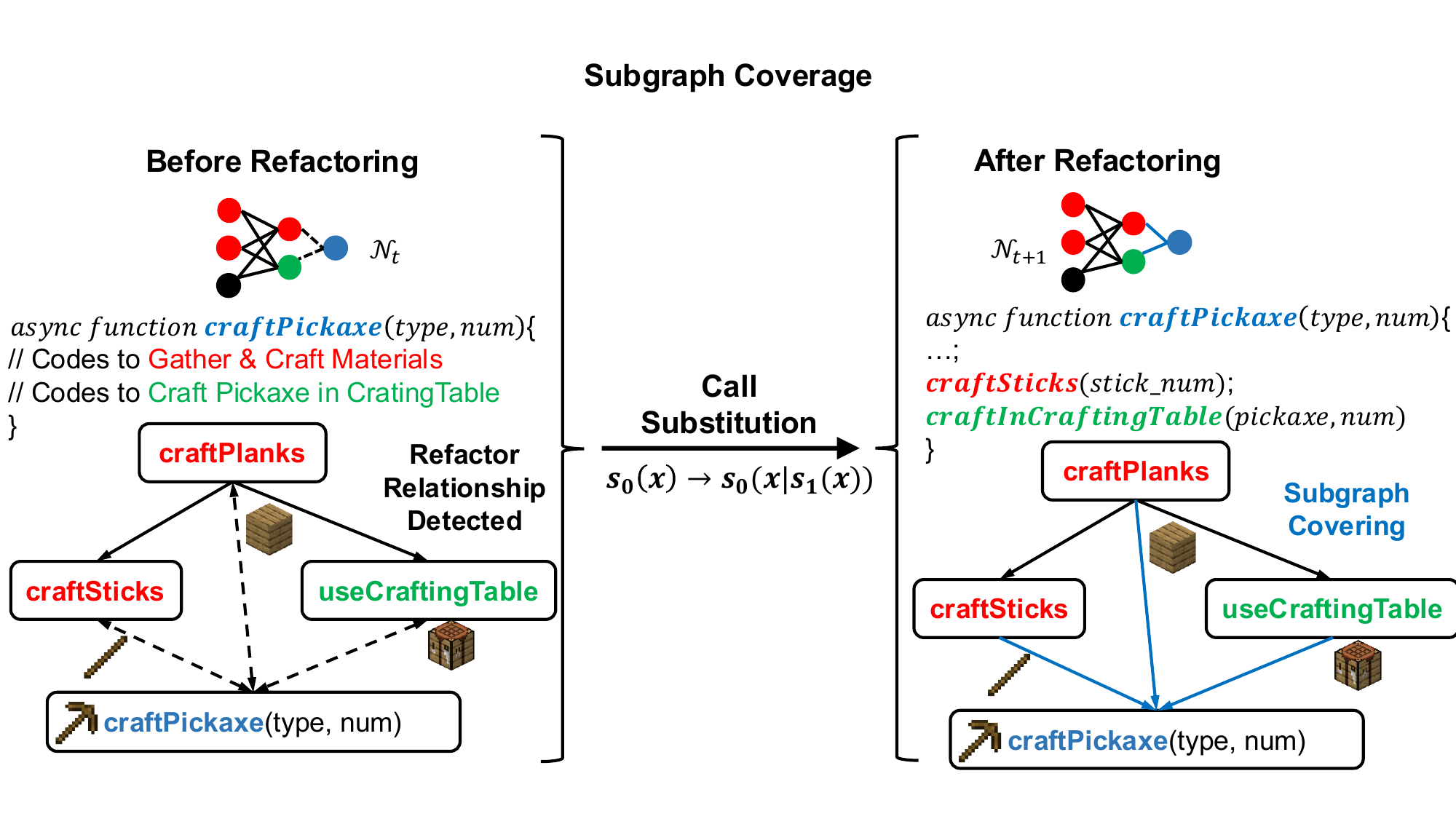}
\caption{Behavioral (subgraph) coverage.
Duplicated logic inside a composite skill is replaced by a call to an existing reusable skill, preserving behavior while reducing redundancy.}
\label{fig:refactor-subgraph}
\end{figure*}

\paragraph{Pattern.}
A composite skill reimplements functionality that already exists as an independent skill in the PSN, resulting in duplicated subgraphs.

\paragraph{Rewrite.}
The duplicated subgraph is removed and replaced by a direct invocation of the existing skill, yielding a simpler and more compositional program structure.

\subsection{Case C: Sibling Specializations}
\label{app:refactor-sibling}

\begin{figure*}[t]
\centering
\includegraphics[width=0.88\textwidth]{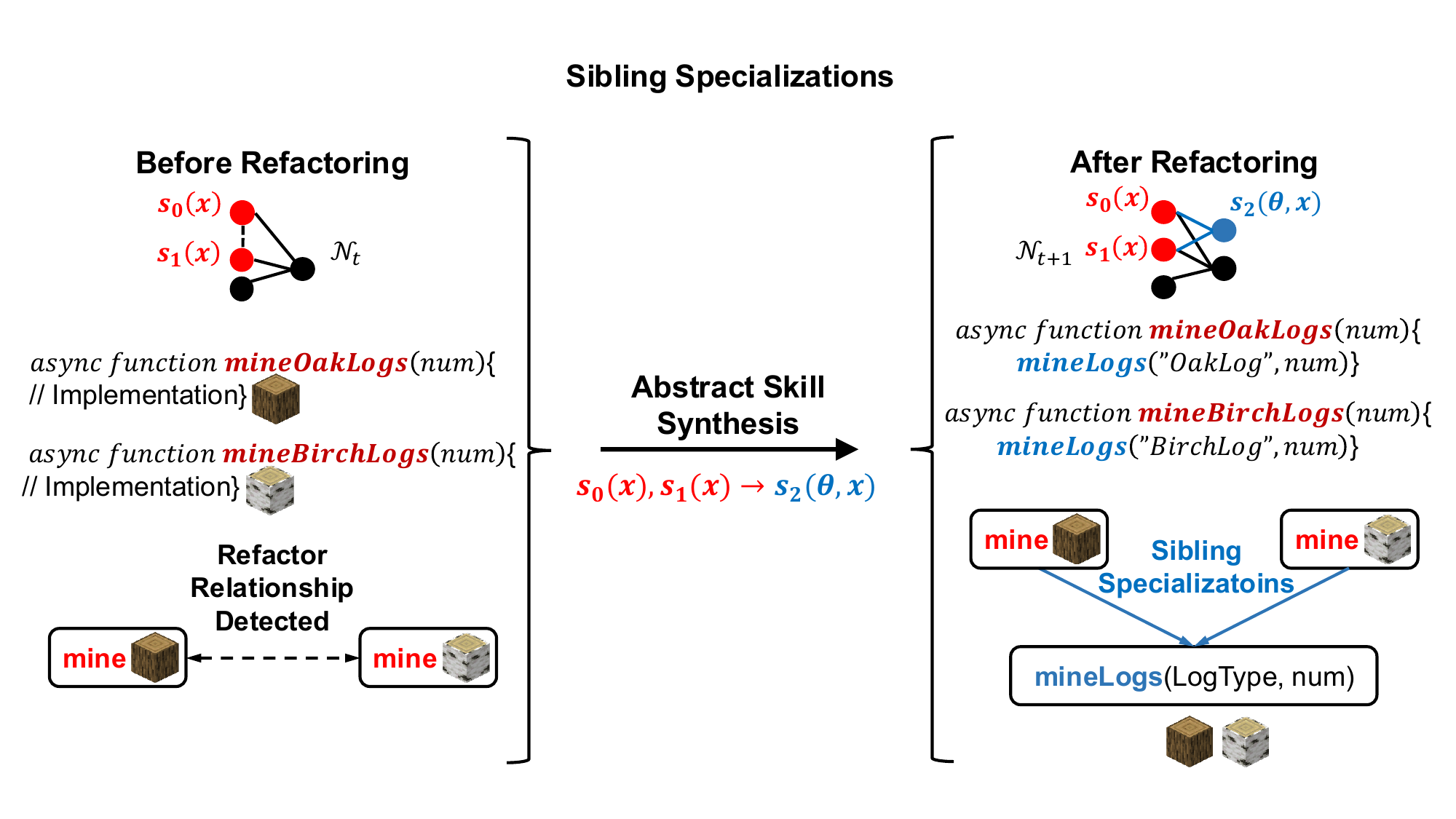}
\caption{Sibling specializations.
Multiple specialized skills expose a missing higher-level abstraction that can be explicitly synthesized and reused.}
\label{fig:refactor-sibling}
\end{figure*}

\paragraph{Pattern.}
Two or more skills are specializations of a latent, more general operation that is not yet represented as a standalone skill in the network.

\paragraph{Rewrite.}
A new abstract skill is synthesized to capture the shared structure, and all specialized skills are rewritten as thin wrappers that invoke the abstract skill with appropriate parameters.

\subsection{Case D: Common Subskill Extraction}
\label{app:refactor-common}

\begin{figure*}[t]
\centering
\includegraphics[width=0.88\textwidth]{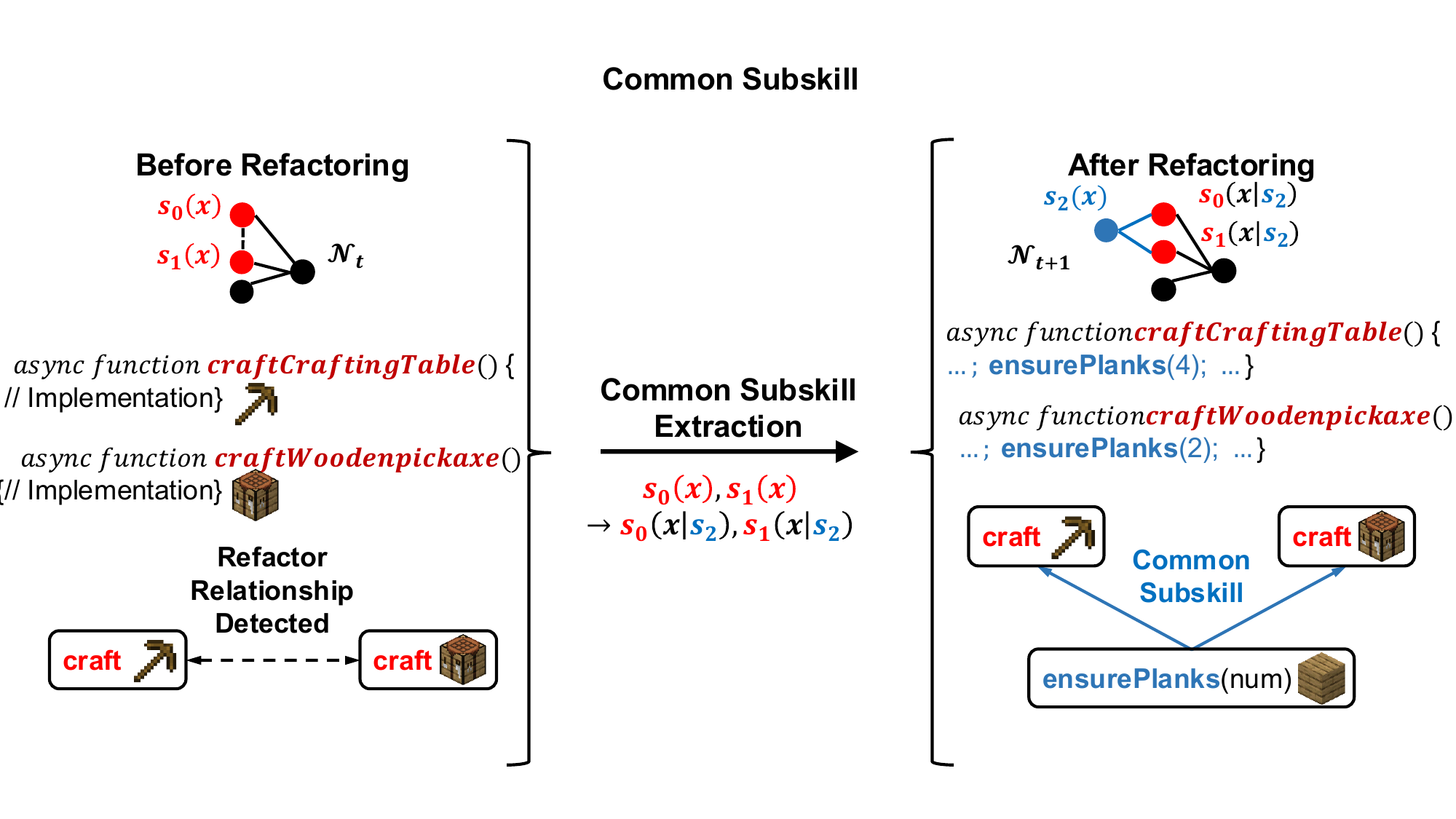}
\caption{Common subskill extraction.
Repeated sub-operations across different skills are factored into a shared subskill, improving reuse and reducing duplication.}
\label{fig:refactor-common}
\end{figure*}

\paragraph{Pattern.}
Multiple skills contain an identical or highly similar sub-operation that is implemented independently within each skill.

\paragraph{Rewrite.}
The shared subgraph is extracted into a new reusable skill, and all original skills are rewritten to invoke this subskill instead of duplicating its logic.

\subsection{Case E: Duplication Removal}
\label{app:refactor-duplication}

\begin{figure*}[t]
\centering
\includegraphics[width=0.88\textwidth]{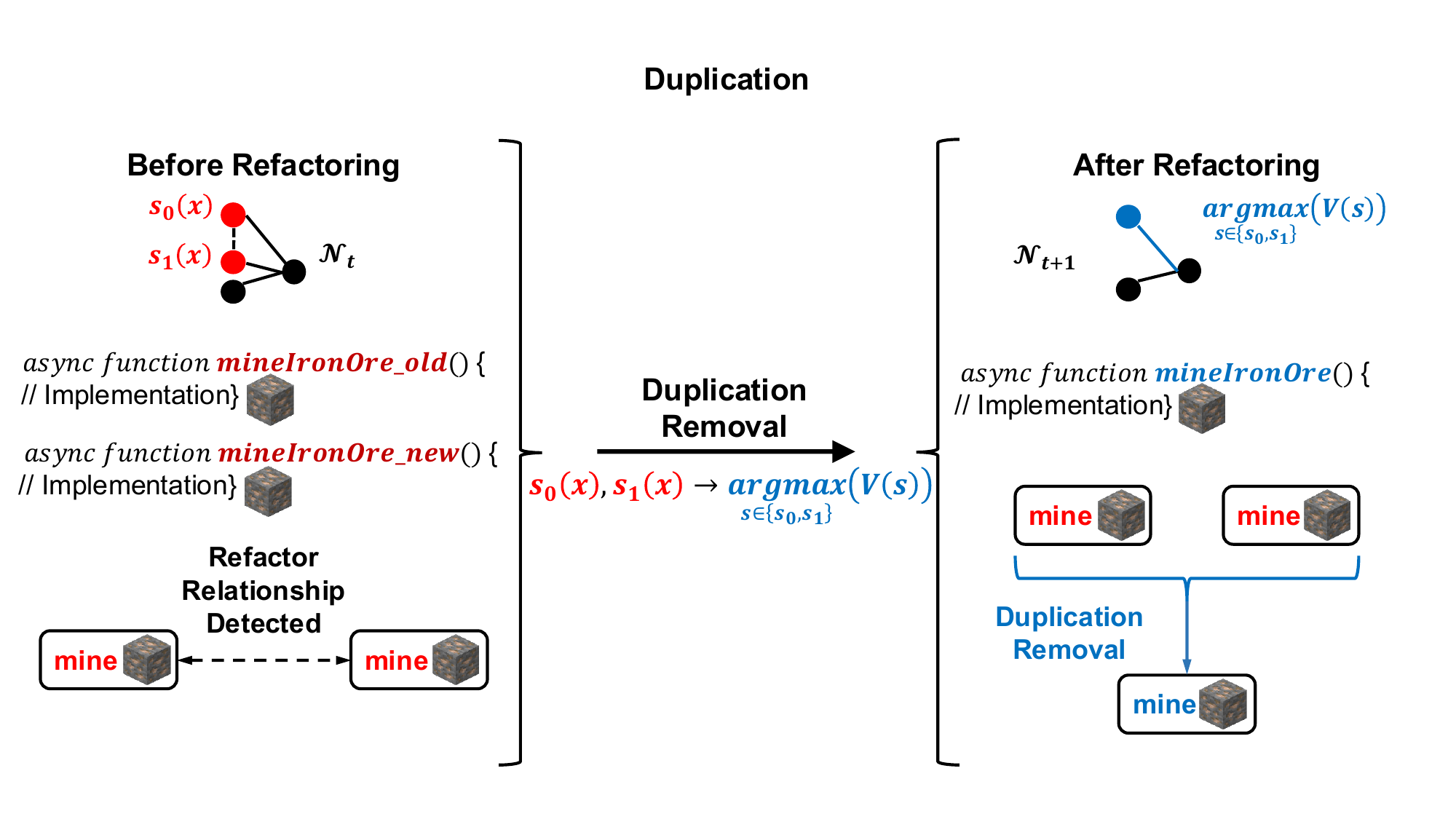}
\caption{Duplication removal. Functionally equivalent skills are merged into a single canonical representation.}
\label{fig:refactor-duplication}
\end{figure*}

\paragraph{Pattern.}
Two skills are functionally equivalent up to naming differences or minor surface variations, leading to redundant representations in the PSN.

\paragraph{Rewrite.}
The skill with higher empirical value is retained as the canonical implementation, and all invocation links to the redundant skill are redirected. The redundant skill is demoted to an alias or removed from planning.

\section{Operator Summary}
\subsection{Symbolic Operators}
\label{app:symbolic-operators}

Table~\ref{tab:symbolic-operators} summarizes the core symbolic operators used in the Programmatic Skill Network (PSN), which define the symbolic forward and backward passes over program-structured skills.

\begin{table*}[t]
\centering
\small
\begin{tabular}{p{0.18\textwidth}p{0.22\textwidth}p{0.27\textwidth}p{0.27\textwidth}}
\hline
\textbf{Operator} & \textbf{Domain $\to$ Codomain} & \textbf{Semantic role} & \textbf{Example in PSN} \\
\hline
$\opexec$ &
$(s, E) \rightarrow (f_s, \delta_s)$ &
Symbolic forward operator that executes a skill program $s$ in environment $E$, producing structured feedback $f_s$ and a success flag $\delta_s \in \{0,1\}$. &
$\opexec(s_{\texttt{craftTable}})$ runs the composed skill
\texttt{craft crafting table}, records the invocation trace and state transitions, and returns whether the goal predicate $g_\tau$ is satisfied. \\[0.5em]
$\opreflect$ &
$(f_s, s) \rightarrow \widetilde{\nabla}_s$ &
Symbolic differentiation operator that performs top-down credit assignment over the PSN, yielding a finite, localized symbolic pseudo-gradient
$\widetilde{\nabla}_s = \partial_s f_s$.
The operator identifies faulty control flow, misaligned parameters, incorrect preconditions, or subskill effects, and serves as a discrete, structural analogue of backpropagation in neural networks. &
$\opreflect(f_{s_{\texttt{craftTable}}}, s_{\texttt{craftTable}})$ detects that \texttt{craftTable} failed due to missing planks and proposes edits to collect wood logs and craft planks for crafting CraftingTable.\\[0.5em]
%
\hline
\end{tabular}
\caption{Symbolic operators defining forward execution and backward credit assignment over program-structured skills in the PSN.}
\label{tab:symbolic-operators}
\end{table*}

\subsection{System Operators}
\label{app:system-operators}

Table~\ref{tab:system-operators} summarizes the system-level operators that orchestrate planning, learning, and structural evolution of the Programmatic Skill Network (PSN).

\begin{table*}[t]
\centering
\small
\begin{tabular}{p{0.18\textwidth}p{0.22\textwidth}p{0.25\textwidth}p{0.27\textwidth}}
\hline
\textbf{Operator} & \textbf{Domain $\to$ Codomain} & \textbf{System role} & \textbf{Example in PSN} \\
\hline
$\opplan$ &
$(g_{\tau_t}, \mathcal{N}_t) \rightarrow P_t^{\text{LLM}}$ &
Fallback forward planner invoked when backward-chaining over existing skills cannot ground a subgoal, producing exploratory plans beyond the current PSN. &
For the task \texttt{``obtain diamond''}, $\opplan$ proposes a long-horizon plan involving mining iron, smelting ingots, crafting pickaxes, and mining diamond ore. \\[0.5em]
$\opcodegen$ &
$(P_t, \text{Context}_t) \rightarrow s_t$ &
Skill synthesis operator that distills a high-level plan into a new symbolic skill neuron with control flow, parameters, and pre/postconditions. &
Given a plan $P_t=[\texttt{getWood},\texttt{craftPlanks},\texttt{craftTable}]$, \opcodegen creates a reusable skill \texttt{craftCraftingTable} with an explicit loop and parameterized inventory checks. \\[0.5em]
$\textsc{Optimize}$ &
$(\mathcal{N}_t, s_t, f_t) \rightarrow \mathcal{N}_{t+1}$ &
Skill optimizer that applies symbolic backpropagation when a task fails, repairing the faulty subnetwork $\mathbb{N}(s_t)$ via \opreflect. &
If \texttt{craftStonePickaxe} fails due to insufficient cobblestone, $\textsc{Optimize}$ propagates symbolic edits to \texttt{mineCobblestone}, inserting a loop until enough stone is collected. \\[0.5em]
$\textsc{Refactor}$ &
$(\mathcal{N}_t, s_t, f_t) \rightarrow \mathcal{N}_{t+1}$ &
Online structural refactor operator that performs symbolic neural architecture search (NAS) when a task succeeds, merging, abstracting, pruning, and rewiring skills. &
After learning both \texttt{mineOakLogs} and \texttt{mineBirchLogs}, $\textsc{Refactor}$ synthesizes a generalized \texttt{mineLogs(log\_type, num)} and rewrites both original skills as wrappers. \\[0.5em]
$\operatorname{embed}$ &
$s \mapsto \operatorname{embed}(s)$ &
Semantic embedding operator used for similarity-based retrieval during refactor, enabling detection of related skills beyond local graph neighborhoods. &
High similarity between $\operatorname{embed}(s_{\texttt{craftStick}})$ and $\operatorname{embed}(s_{\texttt{craftTable}})$ helps identify a common subroutine for ensuring plank availability. \\[0.5em]
$P(\text{update } s)$ &
$V(s) \mapsto [0,1]$ &
Maturity-aware update gate that controls how frequently symbolic derivatives are applied to a skill, stabilizing mature skills while keeping immature ones plastic. &
For a navigation skill with high $V(s)$, $P(\text{update } s)$ becomes small, so \textsc{Optimize} rarely modifies it; newly synthesized skills are updated aggressively until they stabilize. \\
\hline
\end{tabular}
\caption{System-level operators that orchestrate planning, optimization, and structural evolution of the PSN.}
\label{tab:system-operators}
\end{table*}

\section{Example Prompt Templates}
\label{app:prompt_templates}
This appendix provides example prompt templates used to instantiate PSN operators in our implementation. We emphasize that PSN does not rely on specific prompt wording; the examples below serve only as concrete realizations of the abstract operator interfaces defined in Section~\ref{section:method}.

\subsection{REFLECT Operator}
\label{app:reflect_prompt}
The example prompt for REFLECT Operator is demonstrated in Figure~\ref{fig:reflect_prompt}. Note that, to accelerate the speed of REFLECT Operator, we implement an hybrid REFLECT Operator that combine the LLM REFLECT with an rule-based REFLECT function that extract frequent patterns recognized by LLM REFLECT as a set of rules.
\paragraph{Input.}
\begin{itemize}
    \item Skill name and implementation code
    \item Execution feedback and failure signals
    \item Optional execution state, environment context, and child-skill information
\end{itemize}

\paragraph{Output.}
A structured JSON record containing:
\begin{itemize}
    \item Self-responsible issues with gradient type, magnitude, and direction
    \item Child-skill attributions with responsibility weights
    \item Concrete code-level modification suggestions
\end{itemize}

\begin{figure*}[t]
\begin{lstlisting}[language={},basicstyle=\ttfamily\footnotesize,
breaklines=true,columns=fullflexible]

**Skill:** {input.skill_name}

**Code:**
```javascript
{input.skill_code}
```

**Feedback:**
{input.feedback_content}

**Feedback Type:** {input.feedback_type}
{execution_state_section}
{children_section}
{env_section}
{primitive_section}
{propagated_section}
{api_knowledge_section}
{reasoning_examples_section}

**Analysis Tasks:**
1. Identify the root cause of the failure
2. Determine if the issue is in THIS skill or in a child skill
3. For each identified issue, specify:
   - The type of gradient (logic, parameter_semantic, physical_constraint, error_handling, etc.)
   - The magnitude (0.0 to 1.0, higher = more urgent)
   - The direction (what needs to change)
   - The suggested_fix (REQUIRED: concrete code modification suggestions)

**IMPORTANT:** For physical_constraint issues (placement, resource depletion, pathfinding):
- Provide SPECIFIC code changes in suggested_fix
- Example: "Expand maxDistance from 6 to 16, expand vertical search from [-1,1] to [-2,2]"

Return JSON:
{{
    "self_issues": [
        {{
            "gradient_type": "logic|parameter_semantic|physical_constraint|error_handling|interface",
            "magnitude": 0.0-1.0,
            "direction": "what needs to change",
            "evidence": "supporting evidence from feedback",
            "suggested_fix": "REQUIRED: specific code changes to make"
        }}
    ],
    "child_issues": [
        {{
            "child_skill": "name",
            "issue_description": "...",
            "responsibility": "...",
            "weight": 0.0-1.0
        }}
    ],
    "reasoning": "overall analysis"
}}
\end{lstlisting}
\caption{Example prompt template instantiating the \textsc{Reflect} operator.}
\label{fig:reflect_prompt}
\end{figure*}

\subsection{Skill Optimization Operator}
\label{app:opt_prompt}

We instantiate the skill optimization operator as a patching procedure $s \leftarrow \oppatch(s, \tilde{\nabla}_s)$, where $\tilde{\nabla}_s$ is a structured set of issues and modification directions produced by \textsc{Reflect}. The operator consumes a skill implementation together with layered constraints and execution feedback, and outputs a revised implementation along with an explicit requirement-by-requirement audit trail for mandatory fixes. The detailed prompt is demonstrated in Figure~\ref{fig:opt_prompt}.

\begin{figure*}[t]
\centering
\setstretch{0.92}
\begin{lstlisting}[language={},basicstyle=\ttfamily\scriptsize,breaklines=true,columns=fullflexible,keepspaces=true]
=== SYSTEM ===
You are a helpful assistant that optimizes Minecraft skill code.

READ THE LAYERED CONTEXT CAREFULLY!
The context is organized in layers of importance:
- LAYER 1 (MUST FIX): Critical issues that MUST be addressed. Your code will be REJECTED if not fixed.
- LAYER 2 (LOCALIZATION): Specific lines and areas to focus on.
- LAYER 3 (CONSTRAINTS): Rules you must follow (don't change signature, don't redefine external skills).

CRITICAL RULES:
1. Fix ALL issues mentioned in LAYER 1 - these are mandatory
2. Focus your changes on the areas mentioned in LAYER 2
3. Follow ALL constraints in LAYER 3
4. Return COMPLETE code with all brackets matched - do NOT truncate
5. Keep the function signature unchanged
6. Do NOT add new functions with same names as external skills
7. AUTOMATION ONLY - We only support fully automated skills:
   - Use Mineflayer APIs (bot.craft, bot.dig, bot.placeBlock, bot.equip, etc.)
   - Do NOT require user interaction (windowOpen events, "press E", manual operations)
   - Do NOT convert automated code to interactive/manual flows
   - All operations must be programmatic and automatic
8. CODE CONCISENESS: Keep code concise. Do NOT add unnecessary helper functions.
   - Only keep helper functions that are ACTUALLY USED
   - Remove redundant code. If optimized code is longer than original, review and simplify.
9. DO NOT REDEFINE SYSTEM CONTROL PRIMITIVES: The following functions are PROVIDED BY THE SYSTEM.
   DO NOT create local functions with these exact names - they already exist externally:

   mineBlock, craftItem, smeltItem, exploreUntil, placeItem,
   killMob, useChest, givePlacedItemBack, shoot, waitForMobRemoved

   CONTROL PRIMITIVE API SIGNATURES (CRITICAL - Parameter Types):
{primitives_knowledge}

SIMPLIFICATION PRINCIPLE (MANDATORY - Code Bloat Prevention):
{simplification_principle}

ENVIRONMENT KNOWLEDGE AWARENESS:
{environment_knowledge}

Return a JSON object:
{
  "issues": [
    { "type": "issue_type", "description": "brief description" }
  ],
  "optimized_code": "complete optimized code in JavaScript",
  "change_summary": "brief description of changes",
  "requirements_addressed": [
    {
      "requirement_index": 1,
      "how_addressed": "how LAYER 1 requirement was addressed",
      "code_location": "line number or function name"
    }
  ]
}

The "requirements_addressed" field is MANDATORY!
You must explain how EACH requirement from LAYER 1 was addressed.

=== HUMAN ===
Skill: {skill_name}

{edit_context}

FULL CODE (for reference):
{skill_code}
{wrapper_warning}

ADDITIONAL CONTEXT:
Skill description:
{skill_description}

Gradient:
{gradient_summary}

Child skills feedback:
{child_feedback_summary}

{forward_propagation_info}
{current_state_info}

Recent optimization history (last {momentum_window} feedbacks):
{optimization_history}

Statistics:
- Total executions: {total_executions}
- Success rate: {success_rate}
- Failed executions: {failed_executions}

CODE FORMATTING REQUIREMENTS:
- The optimized_code MUST be properly formatted with:
  - One statement per line
  - Proper indentation (2 spaces)
  - Newlines after { and before }
  - DO NOT compress multiple statements into a single line

Return only JSON.
\end{lstlisting}
\caption{Example prompt template instantiating the skill optimization operator
($s \leftarrow \oppatch(s,\tilde{\nabla}_s)$) as a constrained program-repair step.}
\label{fig:opt_prompt}
\end{figure*}

\section{Inline Knowledge Scaffolding Ablation}
\label{app:scaffolding_ablation}

PSN's optimizer Phase~1 prompt builder retrieves and injects inline knowledge from a flat index of $\sim$60 \texttt{KnowledgeItem} dataclass instances covering Mineflayer API behavior (\texttt{api/pathfinder.py}, \texttt{api/bot\_methods.py}, \texttt{api/inventory.py}) and high-level action guidance (\texttt{action\_guidance.py}). Each item has a \texttt{fact} field (the core technical statement) and an optional \texttt{logical\_implications} field listing deductions, diagnostic patterns, or strategy hints. The \texttt{logical\_implications} content is rendered into the REFLECT prompt under a ``Logical Implications:'' header alongside the \texttt{fact}.

This appendix ablates the \texttt{logical\_implications} field to test whether REFLECT's diagnostic capability comes from the architectural mechanism (trace-based credit assignment over execution traces) or from these injected hints.

\paragraph{Methodology.} We use a \emph{stored-prompt A/B} design: instead of running fresh end-to-end training, we replay Phase~1 production prompts (drawn from a previous Minecraft training run) through the live LLM under two conditions:
\begin{itemize}
    \item \textbf{WITH\_SCAFFOLDING}: full Phase~1 prompt with all \texttt{logical\_implications} content rendered.
    \item \textbf{WITHOUT\_SCAFFOLDING}: same prompt with \texttt{logical\_implications} stripped from 52 of 56 items. Three items are retained: one weak-LLM corrigendum (\texttt{tool\_tier\_rules}), one production-tested fix (\texttt{bot.inventory.items}, whose \texttt{.reduce()} pattern guards against a known stacked-items counting bug), and one pure API contract reference (\texttt{checkRecipe\_api\_contract}).
\end{itemize}

Cases include 5 real Phase~1 mining/crafting cases where the actual fix direction is well-known (planted decoys for retrieval-mis-fire bias detection) and 3 synthetic tier-mismatch cases (\texttt{stone\_pickaxe + diamond\_ore}, \texttt{iron\_pickaxe + obsidian}, \texttt{wooden\_pickaxe + iron\_ore}) where the correct diagnosis is ``tool tier insufficient.'' Each (case, condition) pair is sampled repeatedly at temperature=1.0, and we report the fraction of correct diagnoses per condition.

\paragraph{Cross-LLM results.} The \texttt{tool\_tier\_rules} corrigendum has sharply different value on the two models. On Qwen3-Coder-Next, removing it drops diagnostic correctness on the synthetic tier-mismatch cases from 100\% to 67\%, a $+33$pp contribution: the weaker model has genuine Minecraft pretraining gaps, constructing context-anchored hallucinations such as ``lava destroys obsidian'' or ``raw\_iron does not exist in vanilla Minecraft,'' and the corrigendum patches them. On GPT-5-mini the same corrigendum contributes only $+2$pp (the model scores near-100\% with it and 98\% without it): the strong model already knows the tier rules, so the scaffolding is almost inert. The contribution is documented in the codebase via an explicit \texttt{source} field annotation and is independent of the architectural mechanisms.

\paragraph{Baseline cleanliness.} The Qwen3 figure is measured against a cleaned baseline. An earlier baseline left the 52 sibling \texttt{logical\_implications} items in place; several indirectly cued tier-mismatch reasoning, inflating the no-corrigendum accuracy to 83\% and under-reporting the rule's true contribution. Removing those siblings lowers the intrinsic-knowledge baseline to 67\% and reveals the full $+33$pp. GPT-5-mini's baseline is insensitive to this prune (it stays near-100\% either way), so its $+2$pp is unaffected.

\paragraph{Implications.} On the strong-LLM evaluation, REFLECT's diagnostic capability is essentially architectural: removing all inline knowledge scaffolding leaves diagnostic correctness nearly unchanged (a $+2$pp gap). The Qwen3 corrigendum is honest scaffolding for a weaker model with documented Minecraft knowledge gaps, not a hidden algorithmic dependency. This decomposition strengthens the architectural-origin claim of Section~\ref{subsec:psn} by providing direct ablation evidence: the ``rather than prompt-level tricks'' assertion in Section~\ref{subsec:setup} is now verified at the prompt level, not just inferred from end-to-end performance.

\section{Additional Optimization Examples}
\label{sec:appendix-optimizations}

This appendix provides representative examples of execution-level optimizations performed by PSN. All examples are drawn from actual training runs and are selected to illustrate recurring optimization patterns rather than to exhaustively enumerate all repairs. Together, they demonstrate how trace-based symbolic credit assignment enables both localized fixes and coordinated optimization across skill hierarchies. \textbf{Complete code diffs for optimization cases are provided in Section~\ref{app:optimization-diffs}.}

\subsection{Optimization Taxonomy}

Across experiments,  frequent optimizations of PSN fall into several recurring categories.
Table~\ref{tab:optimization-taxonomy} summarizes the most common failure signals and corresponding repair strategies.

\begin{table}[h]
\centering
\small
\begin{tabular}{l l l}
\toprule
Category & Failure Signal & Typical Repair \\
\midrule
Resource miscalculation & insufficient materials & Correct resource accounting \\
Unsafe fallback & silent execution failure & Enforce fail-fast behavior \\
Boundary condition & inventory full & Add capacity-aware constraints \\
Missing preconditions & missing crafting station & Explicit precondition validation \\
API misuse & invalid recipe or action & Correct API invocation \\
Cross-skill contract & downstream semantic failure & Parent--child co-optimization \\
\bottomrule
\end{tabular}
\caption{Common optimization patterns discovered and repaired by PSN.}
\label{tab:optimization-taxonomy}
\end{table}

\subsection{Representative Optimization Cases}

\paragraph{Example 1: Resource Miscalculation (\texttt{craftWoodenPickaxe}).}
\textbf{Failure signal.}
The skill fails during execution with an error indicating insufficient wooden planks.
\textbf{Root cause.}
The original implementation underestimates required resources by ignoring planks consumed during intermediate stick crafting.
\textbf{Repair.}
Using execution traces, PSN localizes the failure to the resource calculation logic and updates the material requirements to account for intermediate crafting steps.
A validation check is added before execution to ensure sufficient materials are available.
\textbf{Outcome.}
After repair, the skill reliably computes correct resource requirements and succeeds across repeated executions.

\paragraph{Example 2: Unsafe Fallback (\texttt{ensureFlint}).}
\textbf{Failure signal.}
The skill exhibits silent or inconsistent failures when attempting to mine gravel.
\textbf{Root cause.}
An unsafe fallback bypasses the system’s primitive execution contract, preventing proper failure propagation to the planner.
\textbf{Repair.}
PSN removes the unsafe fallback and enforces fail-fast behavior, ensuring that execution failures are explicitly surfaced and handled by upstream skills.
\textbf{Outcome.}
The repaired skill behaves consistently and enables reliable replanning under failure.

\paragraph{Example 3: Boundary Condition (\texttt{openChestAndRetrieve}).}
\textbf{Failure signal.}
Execution fails when attempting to retrieve items from a chest due to insufficient inventory capacity.
\textbf{Root cause.}
The skill assumes unlimited inventory space and does not model capacity constraints.
\textbf{Repair.}
The optimizer inserts an explicit capacity check and dynamically constrains the withdrawal amount based on available inventory slots.
\textbf{Outcome.}
The optimized skill adapts to varying inventory states and avoids execution-time errors.

\paragraph{Example 4: Missing Preconditions (\texttt{ensureMetalIngots}).}
\textbf{Failure signal.}
The skill fails when attempting to smelt metal ingots without access to a crafting table or furnace.
\textbf{Root cause.}
The original implementation relies on implicit assumptions about environmental setup.
\textbf{Repair.}
PSN makes these assumptions explicit by validating the presence of required crafting stations and inserting corrective actions to locate or construct them when missing.
\textbf{Outcome.}
The repaired skill succeeds robustly across diverse environment configurations.

\subsection{Advanced Optimization: Cross-Skill Credit Assignment}

Beyond single-skill repairs, PSN is able to propagate optimization signals across skill boundaries.In particular, failures in a parent skill can trigger coordinated updates to both the parent and its dependent subskills.

\paragraph{Example 5: Parent--Child Co-Optimization
(\texttt{ensureRawIronAndFuel} $\rightarrow$ \texttt{ensureFuel}).}
\textbf{Context.}
The parent skill \texttt{ensureRawIronAndFuel} invokes the subskill \texttt{ensureFuel} to acquire sufficient fuel before mining and smelting iron.
\textbf{Failure signal.}
Execution traces show that the parent skill proceeds despite insufficient fuel being present in the inventory, leading to cascading failures in downstream steps.
\textbf{Root cause.}
The parent skill implicitly assumes that successful completion of \texttt{ensureFuel} guarantees the availability of the required fuel.However, the subskill employs coarse fallback behaviors and does not explicitly verify that the desired fuel items are obtained.
\textbf{Coordinated repair.}
PSN assigns credit to both levels of the skill hierarchy and performs simultaneous optimizations:
\begin{itemize}
    \item \textbf{Parent skill repair:} the parent skill is updated to explicitly verify postconditions after invoking the subskill, checking for the presence of coal or charcoal and triggering targeted recovery actions when verification fails.
    \item \textbf{Subskill repair:} the subskill \texttt{ensureFuel} is refined to reduce overly coarse fallbacks, prioritize specific fuel types, and handle inventory-capacity constraints more robustly.
\end{itemize}

\textbf{Outcome.}
After co-optimization, the parent skill reliably enforces its fuel preconditions, and the refined subskill consistently delivers the required resources. This example demonstrates PSN’s ability to localize responsibility across skill boundaries and to perform coordinated, semantics-preserving optimization over compositional skill hierarchies.

%

\definecolor{diffgreen}{RGB}{0, 128, 0}
\definecolor{diffred}{RGB}{180, 0, 0}
\definecolor{diffblue}{RGB}{0, 0, 180}
\definecolor{codebg}{RGB}{250, 250, 250}

\lstdefinelanguage{diff}{
  morecomment=[f][\color{diffblue}]{@@},
  morecomment=[f][\color{diffblue}]{---},
  morecomment=[f][\color{diffblue}]{+++},
  morecomment=[f][\color{diffred}]{-},
  morecomment=[f][\color{diffgreen}]{+},
}

\lstdefinestyle{diffstyle}{
  language=diff,
  basicstyle=\ttfamily\footnotesize,
  backgroundcolor=\color{codebg},
  frame=single,
  framerule=0.5pt,
  breaklines=true,
  breakatwhitespace=false,
  columns=fullflexible,
  keepspaces=true,
  showstringspaces=false,
  numbers=none,
  xleftmargin=3mm,
  xrightmargin=3mm,
  aboveskip=6pt,
  belowskip=6pt,
}

\lstdefinestyle{outputstyle}{
  basicstyle=\ttfamily\footnotesize,
  backgroundcolor=\color{codebg},
  frame=single,
  framerule=0.5pt,
  breaklines=true,
  columns=fullflexible,
  keepspaces=true,
  numbers=none,
  xleftmargin=3mm,
  xrightmargin=3mm,
  aboveskip=4pt,
  belowskip=4pt,
}

\newpage
\onecolumn

\section{Detailed Code Diffs for Optimization Examples}
\label{app:optimization-diffs}

This section provides complete code diffs for the representative optimization cases described in Section~\ref{sec:appendix-optimizations}. Table~\ref{tab:optimization-summary} summarizes all cases, and Table~\ref{tab:gradient-fix-mapping} shows the mapping from gradient signals to implemented fixes.

\begin{table}[h]
\centering
\small
\begin{tabular}{llll}
\toprule
\textbf{Skill} & \textbf{Bug Type} & \textbf{Error Pattern} & \textbf{Key Fix} \\
\midrule
\texttt{craftWoodenPickaxe} & Resource Calc & \texttt{insufficient materials} & Count planks for sticks \\
\texttt{ensureFlint} & Unsafe Fallback & \texttt{Invalid token} & Remove bot.dig() fallback \\
\texttt{openChestAndRetrieve} & Boundary & \texttt{destination full} & Pre-check capacity \\
\texttt{ensureMetalIngots} & Precondition & \texttt{requires crafting table} & Validate \& place table \\
\bottomrule
\end{tabular}
\caption{Summary of optimization cases with bug types and key fixes.}
\label{tab:optimization-summary}
\end{table}

\begin{table}[h]
\centering
\small
\begin{tabular}{p{4.5cm}p{4cm}p{6cm}}
\toprule
\textbf{Gradient Signal} & \textbf{Interpretation} & \textbf{Resulting Fix} \\
\midrule
``Fix resource\_management'' & Math error in counting & Add plank calculation for sticks \\
``fail loudly rather than fallback'' & Unsafe silent failure & Replace fallback with explicit error \\
``Limit withdraw amounts'' & Boundary violation & Add capacity calculation \\
``guarantee crafting table present'' & Missing precondition & Add validation and placement logic \\
\bottomrule
\end{tabular}
\caption{Mapping from gradient signals to implemented fixes.}
\label{tab:gradient-fix-mapping}
\end{table}

\newpage
\subsection{Example 1: craftWoodenPickaxe (Resource Miscalculation)}
\label{app:diff-pickaxe}

\noindent\textbf{Failure Signal.}
\begin{lstlisting}[style=outputstyle]
Error: Cannot craft wooden_pickaxe: insufficient planks. Needed 3, have 0.
\end{lstlisting}

\noindent\textbf{Root Cause.}
The original implementation underestimates required resources by ignoring planks consumed during intermediate stick crafting.

\noindent\textbf{Gradient Signal.}
\begin{lstlisting}[style=outputstyle]
{"gradient_type": "resource_management", "magnitude": 0.9,
 "direction": "Fix plank calculation to include planks consumed by stick crafting"}
\end{lstlisting}

\noindent\textbf{Code Diff.}

\begin{lstlisting}[style=diffstyle]
--- craftWoodenPickaxe.js (original)
+++ craftWoodenPickaxe.js (optimized)
@@ -26,9 +24,19 @@

-    // 2) Ensure we have a crafting_table item in inventory
+    // 2) Ensure crafting_table item by invoking external skill
     bot.chat("No nearby crafting table. Ensuring crafting_table...");
-    await ensureCraftingTable(bot, 1);
+    await ensureCraftingTable(bot, 1, plankType);
+
+    // 2b) Validate ensureCraftingTable result
+    const tableCount = countItemByName("crafting_table");
+    const tableBlock2 = bot.findBlock({
+      matching: mcData.blocksByName["crafting_table"].id,
+      maxDistance: maxDistance
+    });
+    if (tableCount <= 0 && !tableBlock2) {
+      throw new Error("ensureCraftingTable failed.");
+    }
\end{lstlisting}

\begin{lstlisting}[style=diffstyle]
@@ -126,9 +135,11 @@

-    // Totals needed
-    const totalPlanksNeeded = count * planksPerPick;
+    // FIXED: include planks consumed to craft sticks
     const totalSticksNeeded = count * sticksPerPick;
+    const stickRecipesNeeded = Math.ceil(totalSticksNeeded / 4);
+    const planksNeededForSticks = stickRecipesNeeded * 2;
+    const totalPlanksNeeded = (count * planksPerPick) + planksNeededForSticks;
\end{lstlisting}

\begin{lstlisting}[style=diffstyle]
@@ -175,11 +184,44 @@

+    // Recompute planks after crafting sticks
+    havePlanks = countItemByName(plankName);
+    if (havePlanks < totalPlanksNeeded) {
+      const missingPlanks = totalPlanksNeeded - havePlanks;
+      const craftsNeeded2 = Math.ceil(missingPlanks / 4);
+      bot.chat(`Need ${missingPlanks} more ${plankName}.`);
+      for (let j = 0; j < craftsNeeded2; j++) {
+        const check2 = bot.checkRecipe(plankName, 1, null);
+        if (!check2.available) {
+          throw new Error(`Cannot craft ${plankName}: ${check2.message}`);
+        }
+        await bot.craft(check2.recipe, 1, null);
+        await bot.waitForTicks(2);
+      }
+    }

     // 4) Craft wooden_pickaxe at the crafting table
     for (let i = 0; i < count; i++) {
+      // Re-validate resources prior to each craft
+      havePlanks = countItemByName(plankName);
+      haveSticks = countItemByName(stickName);
+      if (havePlanks < planksPerPick) {
+        throw new Error(`Insufficient planks: ${havePlanks}/${planksPerPick}`);
+      }
+      if (haveSticks < sticksPerPick) {
+        throw new Error(`Insufficient sticks: ${haveSticks}/${sticksPerPick}`);
+      }
       const check = bot.checkRecipe("wooden_pickaxe", 1, craftingTableBlock);
\end{lstlisting}

\newpage
\subsection{Example 2: ensureFlint (Unsafe Fallback)}
\label{app:diff-flint}

\noindent\textbf{Failure Signal.}
\begin{lstlisting}[style=outputstyle]
bot.dig failed: Invalid token
Path to gravel failed: Cannot read property 'position' of null
\end{lstlisting}

\noindent\textbf{Root Cause.}
An unsafe fallback using \texttt{bot.dig()} directly bypasses the system's primitive execution contract, preventing proper failure propagation.

\noindent\textbf{Gradient Signal.}
\begin{lstlisting}[style=outputstyle]
{"gradient_type": "error_handling", "magnitude": 0.85,
 "direction": "fail loudly rather than naive fallback"}
\end{lstlisting}

\noindent\textbf{Code Diff.}

\begin{lstlisting}[style=diffstyle]
--- ensureFlint.js (original)
+++ ensureFlint.js (optimized)
@@ -2,13 +2,12 @@

-  // find a nearby gravel block (within 32)
+  // find a nearby gravel block (within SEARCH_RADIUS)
+  const SEARCH_RADIUS = 48;
   function findNearbyGravel() {
     const gravelDef = mcData.blocksByName['gravel'];
     if (!gravelDef) return null;
     return bot.findBlock({
       matching: gravelDef.id,
-      maxDistance: 32
+      maxDistance: SEARCH_RADIUS
     });
   }
\end{lstlisting}

\begin{lstlisting}[style=diffstyle]
@@ -17,61 +16,37 @@

-  // Attempt to mine gravel using mineBlock if available, otherwise fallback
+  // Attempt to mine gravel using mineBlock control primitive
+  // If unavailable, fail loudly so harness can surface the error
   async function mineOneGravelAt(blockPos) {
-    if (typeof mineBlock === "function") {
-      await mineBlock(bot, "gravel", 1);
+    if (typeof mineBlock === 'function') {
+      await mineBlock(bot, 'gravel', 1);
       return;
     }
-    // Fallback manual approach:
-    const targetBlock = bot.blockAt(blockPos);
-    if (!targetBlock) throw new Error("Target gravel disappeared.");
-    try {
-      await bot.pathfinder.goto(new GoalGetToBlock(
-        targetBlock.position.x, targetBlock.position.y, targetBlock.position.z));
-    } catch (e) {
-      bot.chat(`Path to gravel failed: ${e.message}.`);
-    }
-    try {
-      await bot.dig(targetBlock, true);
-    } catch (e) {
-      bot.chat(`dig failed: ${e.message}`);
-      throw e;
-    }
-    await bot.waitForTicks(4);
+    // Deliberately fail if primitive unavailable
+    throw new Error('Required primitive mineBlock is not available.');
   }
\end{lstlisting}

\begin{lstlisting}[style=diffstyle]
@@ -101,10 +78,12 @@

     try {
       await mineOneGravelAt(nearby.position);
     } catch (e) {
+      // Propagate fatal errors for proper handling
       bot.chat(`Failed to mine gravel: ${e.message}`);
+      throw e;
     }
\end{lstlisting}

\newpage
\subsection{Example 3: openChestAndRetrieve (Boundary Condition)}
\label{app:diff-chest}

\noindent\textbf{Failure Signal.}
\begin{lstlisting}[style=outputstyle]
Error: Destination full while withdrawing items from chest
\end{lstlisting}

\noindent\textbf{Root Cause.}
The skill assumes unlimited inventory space and does not model capacity constraints.

\noindent\textbf{Gradient Signal.}
\begin{lstlisting}[style=outputstyle]
{"gradient_type": "physical_constraint", "magnitude": 0.8,
 "direction": "Limit withdraw amounts based on available inventory capacity"}
\end{lstlisting}

\noindent\textbf{Code Diff.}

\begin{lstlisting}[style=diffstyle]
--- openChestAndRetrieve.js (original)
+++ openChestAndRetrieve.js (optimized)
@@ -1,13 +1,38 @@

+  // Helper: compute available inventory space for a specific item
+  function availableInventorySpaceFor(itemId) {
+    const def = mcData.items[itemId] || {};
+    const maxStack = def.stackSize || 64;
+    let free = bot.inventory.emptySlotCount() * maxStack;
+    for (const slot of bot.inventory.items()) {
+      if (slot.type === itemId) {
+        free += (maxStack - slot.count);
+      }
+    }
+    return free;
+  }
+
+  // Helper: get container items across MC versions
+  function getContainerItems(win) {
+    if (!win) return [];
+    if (typeof win.containerItems === 'function') return win.containerItems();
+    if (Array.isArray(win.slots)) return win.slots.filter(Boolean);
+    try { return win.items(); } catch (e) { return []; }
+  }
\end{lstlisting}

\begin{lstlisting}[style=diffstyle]
@@ -65,56 +88,81 @@

+        // Compute how many items are actually in the chest
+        const availableInChest = containerItems
+          .filter(i => i && i.type === itemDef.id)
+          .reduce((s, i) => s + (i.count || 0), 0);
+
+        const capacity = availableInventorySpaceFor(itemDef.id);
+        const toWithdraw = Math.min(want, availableInChest, capacity);
+
+        if (toWithdraw <= 0) {
+          await bot.chat(`No space or chest lacks ${name}, skipping.`);
+          withdrawn[name] = bot.inventory.count(itemDef.id, null);
+          continue;
+        }

         try {
-          await chestWindow.withdraw(itemDef.id, null, want);
+          await chestWindow.withdraw(itemDef.id, null, toWithdraw);
           await bot.waitForTicks(3);
+          withdrawn[name] = bot.inventory.count(itemDef.id, null);
         } catch (err) {
+          if (err.message.includes('destination full')) {
+            const controlled = new Error('Destination full');
+            controlled.code = 'DESTINATION_FULL';
+            throw controlled;
+          }
           throw err;
         }
\end{lstlisting}

\newpage
\subsection{Example 4: ensureMetalIngots (Missing Precondition)}
\label{app:diff-metal}

\noindent\textbf{Failure Signal.}
\begin{lstlisting}[style=outputstyle]
Error: No furnace recipe available (missing materials).
Error: Failed to find or place a crafting table before crafting furnace.
\end{lstlisting}

\noindent\textbf{Root Cause.}
The original implementation relies on implicit assumptions about environmental setup without validating the presence of required crafting stations.

\noindent\textbf{Gradient Signal.}
\begin{lstlisting}[style=outputstyle]
{"gradient_type": "precondition", "magnitude": 0.9,
 "direction": "guarantee crafting table is present before furnace craft"}
\end{lstlisting}

\noindent\textbf{Code Diff.}

\begin{lstlisting}[style=diffstyle]
--- ensureMetalIngots.js (original)
+++ ensureMetalIngots.js (optimized)
@@ -68,81 +70,144 @@

+      // If no placed table and no table item, craft one (2x2 recipe)
+      if (!craftingBlock && !craftingItemInv && craftingItemDef) {
+        try {
+          const tableRecipes = bot.recipesFor(craftingItemDef.id, null, null) || [];
+          if (tableRecipes.length > 0) {
+            await bot.craft(tableRecipes[0], 1, null);
+            await bot.waitForTicks(4);
+            craftingItemInv = bot.inventory.findInventoryItem(craftingItemDef.id);
+          }
+        } catch (e) {
+          bot.chat(`Crafting crafting_table failed: ${e.message}`);
+        }
+      }
+
+      // If we have table item but no placed block, place it
+      if (!craftingBlock && craftingItemInv) {
+        const botFoot = bot.entity.position.floored();
+        const searchOffsets = [
+          new Vec3(1, 0, 0), new Vec3(-1, 0, 0),
+          new Vec3(0, 0, 1), new Vec3(0, 0, -1),
+        ];
+        let candidate = null;
+        for (const off of searchOffsets) {
+          const cand = botFoot.offset(off.x, off.y, off.z);
+          if (cand.equals(botFoot)) continue;
+          candidate = cand;
+          break;
+        }
+
+        try {
+          await bot.pathfinder.goto(new GoalPlaceBlock(candidate, bot.world, {}));
+        } catch (e) {
+          bot.chat(`Path to table spot failed: ${e.message}`);
+        }
+
+        await bot.equip(craftingItemInv, "hand");
+        await bot.placeBlock(ref, new Vec3(0, 1, 0));
+        await bot.waitForTicks(4);
+        craftingBlock = bot.blockAt(candidate);
+
+        // Verify placement succeeded
+        if (!craftingBlock || craftingBlock.name !== "crafting_table") {
+          throw new Error("Crafting table placement failed.");
+        }
+      }
+
+      // Ensure placed crafting table before furnace craft
+      if (!craftingBlock) {
+        throw new Error("Failed to find or place a crafting table.");
+      }
+
+      // Move within interaction distance
+      try {
+        await bot.pathfinder.goto(new GoalNear(
+          craftingBlock.position.x, craftingBlock.position.y,
+          craftingBlock.position.z, 2));
+      } catch (e) {
+        bot.chat(`Could not move near crafting table: ${e.message}`);
+      }
+
+      // Use recipesFor instead of checkRecipe
+      if (!furnaceItemDef) throw new Error('Furnace definition missing.');
+      const furnaceRecipes = bot.recipesFor(furnaceItemDef.id, null, craftingBlock) || [];
+      if (furnaceRecipes.length === 0) {
+        throw new Error('No furnace recipe available.');
+      }
+      try {
+        await bot.craft(furnaceRecipes[0], 1, craftingBlock);
+        await bot.waitForTicks(4);
+      } catch (e) {
+        throw new Error(`Crafting furnace failed: ${e.message}`);
+      }
\end{lstlisting}

\newpage
\subsection{Example 5: Cross-Skill Co-Optimization}
\label{app:diff-cross-skill}

Beyond single-skill repairs, PSN propagates optimization signals across skill boundaries. This example shows coordinated parent--child optimization between \texttt{ensureRawIronAndFuel} (parent) and \texttt{ensureFuel} (child).

\paragraph{Failure Signal.}
The parent skill proceeds despite insufficient fuel, causing cascading failures in downstream smelting steps.

\noindent\textbf{Coordinated Repair.}
PSN assigns credit to both levels of the hierarchy and performs simultaneous optimizations.

\vspace{0.5em}
\noindent\textit{Parent skill repair} (\texttt{ensureRawIronAndFuel}):

\begin{lstlisting}[style=diffstyle]
+    // Verify fuel postcondition after calling ensureFuel
+    const fuelCount = countItemByName("coal") + countItemByName("charcoal");
+    if (fuelCount < requiredFuel) {
+      bot.chat(`ensureFuel insufficient: ${fuelCount}/${requiredFuel}`);
+      await ensureFuel(bot, requiredFuel - fuelCount, "coal");
+    }
\end{lstlisting}

\vspace{0.5em}
\noindent\textit{Child skill repair} (\texttt{ensureFuel}):

\begin{lstlisting}[style=diffstyle]
-    const fuels = ["coal", "charcoal", "oak_log", "birch_log"];
+    // Prioritize efficient fuel sources
+    const fuels = preferredFuel 
+      ? [preferredFuel, "coal", "charcoal"] 
+      : ["coal", "charcoal"];
     for (const fuel of fuels) {
-      if (tryGetFuel(fuel)) return;
+      const obtained = await tryGetFuel(fuel, needed - currentFuel);
+      currentFuel += obtained;
+      if (currentFuel >= needed) break;
     }
+    // Explicit postcondition check
+    if (currentFuel < needed) {
+      throw new Error(`ensureFuel failed: ${currentFuel}/${needed}`);
+    }
\end{lstlisting}

This demonstrates PSN's ability to localize responsibility across skill boundaries and perform coordinated optimization over compositional skill hierarchies.

\section{Task Sequences}
\label{app:task_sequences}

\paragraph{Temporal Generalization Curriculum (Section~\ref{subsec:setup}).}
\textit{Mine wood} $\rightarrow$ \textit{Craft table} $\rightarrow$ \textit{Craft wooden pickaxe} $\rightarrow$ \textit{Craft stone pickaxe} $\rightarrow$ \textit{Mine iron} $\rightarrow$ \textit{Smelt iron} $\rightarrow$ \textit{Craft iron pickaxe}.

\paragraph{Offline vs.\ Online Refactor Evaluation (Section~4.4).}
\textit{Mine wood} $\rightarrow$ \textit{Craft planks} $\rightarrow$ \textit{Craft table} $\rightarrow$ \textit{Craft wooden pickaxe} $\rightarrow$ \textit{Mine cobblestone} $\rightarrow$ \textit{Craft stone pickaxe} $\rightarrow$ \textit{Mine iron} $\rightarrow$ \textit{Smelt iron} $\rightarrow$ \textit{Craft iron pickaxe}.
All methods are evaluated on the identical task sequence without retraining.

\section{Comparison with ODYSSEY}
\label{app:odyssey_comparison}

\paragraph{Autonomy--Engineering Spectrum.}
PSN, ADAM, and ODYSSEY represent three positions along an autonomy--engineering spectrum. ODYSSEY represents the \emph{maximum-engineering} endpoint: 183 hand-authored JavaScript skills achieve deterministic first-attempt success but require extensive per-domain human programming and cannot adapt beyond their pre-built repertoire. ADAM occupies an intermediate position, relying on 47 hand-authored skills while learning their causal input-output mappings via intervention-based verification. PSN represents the \emph{maximum-autonomy} endpoint: without hand-authored task-specific skills, it generates all skill code from scratch, reaching diamond tool in 35$\pm$16 iterations (6/6 runs) with 30 autonomously learned skills---fewer iterations and higher reliability than ADAM (74$\pm$23) and Voyager* (99$\pm$36, 2/6). Table~\ref{tab:systematic_comparison} provides a systematic comparison.

\begin{table*}[h]
\small
\centering
\resizebox{\textwidth}{!}{%
\begin{tabular}{l|l|l|l}
\toprule
\textbf{Dimension} & \textbf{PSN (Ours)} & \textbf{ODYSSEY} & \textbf{ADAM} \\
\midrule
Skill Source & LLM-generated \& optimized (\textbf{30} learned) & \textbf{183} hand-authored JS skills & 47 hand-authored JS skills + learned causal I/O mappings \\
Per-Task Skill Engineering & \textbf{None} & High (manual JS authoring per task domain) & High (47 JS skills + unlock tree + item/action dictionaries) \\
Skill Evolution & Yes (\opreflect + maturity gating) & No (static, fixed at authoring time) & No (causal graph is static once learned) \\
Code-Level Optimization & Yes (trace-based fault localization) & N/A (pre-built code) & N/A (no code generation) \\
Failure Recovery & Trace-based iterative repair & N/A (pre-built) & Memoryless restart (no cross-attempt accumulation) \\
Learning Capability & Generates, composes, optimizes code & None (retrieval of pre-built skills) & Shallow (discovers precondition-effect pairs only) \\
Tech Tree Completion & 100\% (6/6, both LLMs) & 100\% (pre-built skills, 120 ep.) & 100\% (3/3, GPT-5-mini; 0/3 Qwen3) \\
Diamond Tool & 35$\pm$16 iter (6/6) / 49$\pm$18 (Qwen3) & Not reported & Not reported \\
\bottomrule
\end{tabular}%
}
\caption{Systematic comparison of PSN, ODYSSEY~\citep{liu2024odyssey}, and ADAM~\citep{zhang2024adam}. ODYSSEY and ADAM both rely on hand-authored task-specific skills; PSN learns all skills autonomously from scratch.}
\label{tab:systematic_comparison}
\end{table*}

\paragraph{Learning Trajectory: PSN vs.\ ODYSSEY (SR@k).}
To illustrate PSN's self-improvement capability, we evaluate task success rate at $k$ attempts (SR@$k$) on the same 6-subgoal tech tree (Table~\ref{tab:srk}).

\begin{table}[h]
\small
\centering
\begin{tabular}{l|l|cccc}
\toprule
\textbf{Method} & \textbf{Skills} & \textbf{SR@1} & \textbf{SR@2} & \textbf{SR@3} & \textbf{SR@4} \\
\midrule
ODYSSEY & 183 pre-built & 100\% & 100\% & 100\% & 100\% \\
\textbf{PSN} & \textbf{30 learned} & 67\% & 83\% & \textbf{100\%} & \textbf{100\%} \\
\bottomrule
\end{tabular}
\caption{Success rate at $k$ attempts (SR@$k$) on the Minecraft tech tree. Both methods use GPT-5-mini.}
\label{tab:srk}
\end{table}

ODYSSEY's pre-built skills achieve deterministic first-attempt success, as expected for hand-engineered, pre-tested code. Notably, when we load ODYSSEY's 183 skills into our evaluation framework with Qwen3-Coder for planning, the system achieves 100\% SR@1 across all 18 subgoal evaluations (3 trials $\times$ 6 subgoals) \emph{without any optimization or adaptation}. This confirms that ODYSSEY's improvements stem from the quality of its hand-authored skill library rather than from algorithmic learning.

By contrast, PSN starts with lower SR@1 (skills are learned from scratch, not pre-built) but converges to 100\% at SR@3 through autonomous trace-based optimization (\opreflect). This SR@1$\to$SR@3 progression directly demonstrates PSN's self-improvement capability (the core contribution of this work) is achieved \textbf{without hand-authored task-specific skills}. PSN achieves this with only 30 autonomously learned skills, compared to ODYSSEY's 183 hand-authored skills, which is a 6$\times$ reduction in skill count while matching final reliability, highlighting the compositional efficiency of PSN's learned skill network.

\section{Credit Assignment Analysis}
\label{app:credit_assignment}

This appendix provides a quantitative analysis of PSN's credit assignment mechanism, examining how trace-based fault localization propagates repairs across skill compositions.

\subsection{Per-Episode Optimization Scope}

We measure the \emph{actual scope} of each optimization episode: when a task fails, how many skills in the invocation trace does PSN repair? An episode groups all skill optimizations triggered by a single task failure. A \emph{multi-skill episode} indicates that \textsc{Reflect}'s recursive trace analysis propagated repairs across multiple skills in the composition.

\begin{table}[h]
\small
\centering
\begin{tabular}{lcccc}
\toprule
\textbf{Model} & \textbf{Episodes} & \textbf{Single-skill} & \textbf{Multi-skill} & \textbf{Avg depth} \\
\midrule
GPT-5-mini & 204 & 88 (43.1\%) & 116 (56.9\%) & 2.7 \\
Qwen3-Coder-Next & 78 & 15 (19.2\%) & 63 (80.8\%) & 5.0 \\
\bottomrule
\end{tabular}
\caption{Per-episode optimization scope across all experimental runs. Multi-skill episodes indicate that PSN's trace-based credit assignment propagated repairs across multiple skills in a composition. Qwen3-Coder-Next exhibits deeper credit propagation (avg 5.0 skills per multi-skill episode) than GPT-5-mini (2.7).}
\label{tab:attribution_distribution}
\end{table}

Both models exhibit substantial cross-skill credit assignment: 56.9\% (GPT-5-mini) and 80.8\% (Qwen3-Coder-Next) of optimization episodes involve coordinated repairs across multiple skills in a composition. Qwen3 shows deeper propagation (average 5.0 skills per multi-skill episode vs.\ 2.7 for GPT-5-mini), reflecting that weaker code generators produce compositions requiring more extensive repair chains. Crucially, this cross-skill optimization occurs through PSN's recursive trace-based fault localization, which is the architectural mechanism, rather than the LLM's explicit attribution judgment, drives effective multi-skill repair.

\subsection{Error Mode Taxonomy}
\label{app:error_modes}

We categorize the failure modes identified by \textsc{Reflect} across 1{,}059 optimization records from all experimental runs. Since a single optimization record may exhibit multiple failure modes, percentages sum to greater than 100\%.

\begin{table}[h]
\small
\centering
\begin{tabular}{lrl}
\toprule
\textbf{Error Mode} & \textbf{\%} & \textbf{Typical Scope} \\
\midrule
Resource miscalculation & 49.9\% & Single-skill \\
API contract violation & 27.8\% & Single-skill \\
Precondition gap & 24.3\% & Single-skill \\
Defensive validation & 23.2\% & Single-skill \\
Placement / movement & 21.0\% & Single-skill \\
Return value contract & 8.8\% & Multi-skill \\
Cross-skill postcondition & 6.1\% & Multi-skill \\
Parameter / interface mismatch & 2.2\% & Multi-skill \\
LLM hallucination & 0.3\% & Single-skill \\
\bottomrule
\end{tabular}
\caption{Error mode distribution across 1{,}059 optimization records. Single-skill errors (top 5 rows) are resolved within individual skills; inter-skill contract violations (rows 6--8) trigger multi-skill repair episodes where \textsc{Reflect} propagates fixes across the invocation trace. $N = 1{,}059$; percentages sum to $>$100\% due to multi-label classification.}
\label{tab:error_modes}
\end{table}

Resource miscalculation dominates (49.9\%), consistent with Minecraft's deep material dependency chains where intermediate crafting steps consume resources that must be accounted for. Cross-boundary error modes, including return value contracts (8.8\%), cross-skill postconditions (6.1\%), and parameter mismatches (2.2\%), collectively account for $\sim$17\% of failures and typically trigger multi-skill repair episodes, confirming that \textsc{Reflect}'s recursive trace analysis identifies and propagates repairs for inter-skill issues.

Notably, LLM hallucination (generating references to non-existent functions or APIs) is extremely rare (0.3\%, 3 cases across all runs). PSN's structured prompting and validation effectively alleviates the LLM's hallucination in response.

\subsection{Structural Credit Assignment}

The per-episode analysis reveals that PSN's credit assignment operates at the \emph{system} level rather than the \emph{judgment} level. When a task fails, \textsc{Reflect} recursively analyzes the invocation trace top-down, identifying faulty skills at each level. Each skill is then optimized bottom-up. This recursive structure naturally propagates repairs across the skill graph without requiring the LLM to explicitly categorize faults as ``caller'' or ``callee'' issues---the compositional credit assignment emerges from the trace-based architecture itself. This is evidenced by the high multi-skill episode rates for both GPT-5-mini (56.9\%) and Qwen3-Coder-Next (80.8\%), demonstrating that the architectural mechanism generalizes across LLM backends of varying capability.

\section{Rollback and Safety Statistics}
\label{app:rollback_stats}

PSN employs a two-stage safety pipeline for structural changes to the skill network:

\begin{enumerate}[leftmargin=1.5em]
\item \textbf{Pre-application validation} (Stage 1): Before any code is modified, each refactor proposal undergoes syntax checking, type safety verification, and semantic preservation analysis. Proposals that fail these checks are rejected outright, i.e., no skill code is changed. The rejection rate for this stage is reported in the refactoring case distribution (Table~\ref{tab:refactor_dist}: 43 of 139 proposals rejected, 31\%).

\item \textbf{Post-application performance rollback} (Stage 2): Proposals that pass Stage 1 validation are applied to the skill network. The system then evaluates performance on a sliding window of 3 recent tasks involving the affected skills. If the success rate drops by more than 20\%, the refactor is reverted using logged inverse operations. Table~\ref{tab:rollback_stats} reports these post-application rollback events.
\end{enumerate}

Post-application rollbacks (Table~\ref{tab:rollback_stats}) are rare in both model settings (3.1\% and 6.7\% of iterations, respectively), because Stage 1 validation already filters most problematic proposals. The weaker open-source model experiences approximately double the rollback rate, with both performance degradation rollbacks and refactor failure rollbacks increasing substantially. This suggests that PSN's safety mechanisms scale with model capability: weaker models produce more optimization regressions that the degradation monitor catches, as well as more refactoring errors that the rollback procedure handles.

\begin{table}[h]
\small
\centering
\begin{tabular}{lccccc}
\toprule
\textbf{Model} & \textbf{Iterations} & \textbf{Rollbacks} & \textbf{Rate} & \textbf{Perf.\ Degrad.} & \textbf{Refactor Fail} \\
\midrule
GPT-5-mini & 131 & 4 & 3.1\% & 1 & 3 \\
Qwen3-Coder-Next & 180 & 12 & 6.7\% & 6 & 6 \\
\bottomrule
\end{tabular}
\caption{Rollback statistics across model settings. ``Perf.\ Degrad.'' indicates rollbacks triggered by success rate drops exceeding the 20\% threshold on a sliding window of recent tasks. ``Refactor Fail'' indicates rollbacks due to refactoring errors (syntax, type, or semantic validation failures detected post-application).}
\label{tab:rollback_stats}
\end{table}

This reflects a deliberate safety-flexibility tradeoff. The five canonical refactoring patterns (Table~\ref{tab:refactor-casebook}) are chosen to be semantics-preserving, each admitting a deterministic rewrite rule. The refactoring framework is modular: new canonical patterns can be added by defining (1)~a detection criterion, (2)~a deterministic rewrite rule, and (3)~a rollback procedure. Extending the pattern set to cover additional structural relationships is a natural direction for future work.

\section{Refactoring Case Distribution}
\label{app:refactor_distribution}

Table~\ref{tab:refactor_dist} reports the aggregate distribution of refactoring cases across all experimental runs with refactoring enabled.

\begin{table}[h]
\small
\centering
\begin{tabular}{lcc}
\toprule
\textbf{Refactor Type} & \textbf{Count} & \textbf{\%} \\
\midrule
Behavioral coverage & 60 & 43\% \\
Extract common subskill & 48 & 35\% \\
Parametric coverage & 18 & 13\% \\
Merge siblings & 11 & 8\% \\
Duplication removal & 2 & 1\% \\
\midrule
\textbf{Total proposed} & \textbf{139} & \textbf{100\%} \\
\textbf{Successfully applied} & \textbf{96} & \textbf{69\%} \\
\bottomrule
\end{tabular}
\caption{Distribution of canonical refactoring cases. 96 of 139 proposed refactors (69\%) pass pre-application validation and are applied; the remaining 31\% are rejected by syntax, type safety, or semantic preservation checks before any code is modified.}
\label{tab:refactor_dist}
\end{table}

Behavioral coverage and common subskill extraction dominate (43\% and 35\%, respectively), consistent with the organic growth pattern of a skill network: as new skills are synthesized, they frequently reimplement functionality that already exists (behavioral coverage) or share common sub-operations (common subskill extraction). Duplication removal is rare (1\%) because earlier refactoring patterns catch overlap before it reaches full duplication. The 31\% rejection rate by pre-application validation confirms that the safety mechanism is binding, preventing potentially harmful structural changes from being applied to the skill network.

\section{API Cost and Token Consumption}
\label{app:api_cost}

Table~\ref{tab:api_cost} compares per-task token consumption and total cost between PSN and Voyager.

\begin{table}[h]
\small
\centering
\begin{tabular}{lcc}
\toprule
\textbf{Metric} & \textbf{PSN} & \textbf{Voyager*} \\
\midrule
Avg.\ tokens per task & $\sim$107K & $\sim$30K \\
Overhead ratio & \multicolumn{2}{c}{3.5$\times$} \\
Total cost to Diamond Tool & $\sim$\$1.76 (3/3 runs) & $\infty$ (0/3 runs) \\
Most expensive run & \$2.27 & -- \\
\bottomrule
\end{tabular}
\caption{Token consumption and cost comparison. PSN incurs $\sim$3.5$\times$ per-task overhead but reaches Diamond Tool in all runs; Voyager* never reaches Diamond Tool. Costs computed at GPT-5-mini pricing.}
\label{tab:api_cost}
\end{table}

Comparing per-task token consumption directly is not a fair comparison, because PSN deliberately invests more tokens in early stages to learn, optimize, and compose a high-quality skill network. This early investment is precisely what enables efficient skill reuse in later stages, unlocking harder achievements and sustaining longer survival. If we only look at per-task cost in the early tech tree, PSN appears more expensive; but PSN reaches Diamond Tool in all six runs while Voyager* does so in only 2/6 runs.

This investment amortizes clearly over the run. We observe \opcodegen overhead dropping from 8.3 calls per task in the first third of training to 5.0 in the final third (a 40\% reduction), because later tasks increasingly compose existing skills rather than generating code from scratch. Once optimized, skills incur zero marginal LLM cost at execution time. For instance, \texttt{ensureCraftingTable} (reused 11 times in one run) executes without any LLM call after initial optimization, and the planning phase composes existing skills via graph traversal rather than LLM generation.

\section{Minecraft Agent Paradigm Taxonomy}
\label{app:taxonomy}

Table~\ref{tab:taxonomy} categorizes Minecraft LLM agents by their skill representation, environment backend, action space, learning paradigm, and primary evaluation metric. This taxonomy clarifies why only code-generating, Mineflayer-based agents support controlled comparison with PSN.

\begin{table*}[h]
\small
\centering
\resizebox{\textwidth}{!}{%
\begin{tabular}{llllllc}
\toprule
\textbf{Method} & \textbf{Venue} & \textbf{Skill Representation} & \textbf{Env Backend} & \textbf{Action Space} & \textbf{Primary Metric} & \textbf{Comparable?} \\
\midrule
Voyager & TMLR'24 & LLM-generated JS code & Mineflayer & Code generation & Iterations to unlock & \checkmark \\
ADAM & ICLR'25 & 47 hand-authored JS skills & Mineflayer & Predefined action selection & Iterations to unlock & \checkmark \\
ODYSSEY & IJCAI'25 & 183 hand-authored JS skills & Mineflayer & Code generation & SR@$k$ & \checkmark \\
\midrule
DEPS & NeurIPS'23 & Predefined goal-cond.\ policies & MC-Controller & Goal-conditioned & Success rate & -- \\
JARVIS-1 & T-PAMI'24 & Predefined MineRL actions & MineRL / MCP & Keypresses & Success rate & -- \\
RL-GPT & NeurIPS'24 & RL policy & MineDojo & Keyboard / mouse & Reward & -- \\
Optimus-1 & NeurIPS'24 & VLM + Steve-1 controller & MineRL & Vision policy & Success rate & -- \\
ROCKET-1 & arXiv'24 & Visual-motor policy & MineDojo & Segmentation actions & Reward & -- \\
\bottomrule
\end{tabular}%
}
\caption{Taxonomy of Minecraft LLM agent paradigms. ``Comparable?'' indicates whether direct experimental comparison with PSN is methodologically sound, requiring the same environment backend (Mineflayer), compatible action space (code generation), and comparable evaluation metrics. Methods below the mid-rule operate in fundamentally different paradigms.}
\label{tab:taxonomy}
\end{table*}

\paragraph{Non-Minecraft programmatic skill methods.}
Several recent works share PSN's interest in programmatic skill/tool learning but operate in fundamentally different domains, precluding controlled comparison. ReGAL~\citep{stengel-eskin2024regal} performs \emph{offline} refactoring over static code benchmarks (including TextCraft, a text-only symbolic crafting simulator with no 3D world or real-time physics). WALT~\citep{prabhu2025walt} discovers browser automation tools via offline demonstration-generation-validation cycles in web DOM environments. ASI~\citep{wang2025asi} induces short browser action scripts (2--5 steps) online during web task solving, which is the closest to PSN's online paradigm, but targeting flat browser primitives rather than hierarchical, long-horizon skill compositions in a 3D embodied world. These methods share methodological concepts (programmatic abstractions, skill reuse) but differ in environment (text/web vs.\ embodied 3D), action complexity (browser macros vs.\ async multi-step game programs), and evaluation infrastructure, making direct numerical comparison infeasible.  

\section{Learning Dynamics Decomposition}
\label{app:jn_decomposition}

We empirically track the four components of the composite objective $J(\mathcal{N})$ introduced in Section~\ref{sec:optimization} across training iterations. This section defines how each component is computed, then analyzes the resulting learning trajectories.

\subsection{Component Definitions}

Each component is a scalar in $[0, 1]$ computed from the current network state:

\begin{itemize}[leftmargin=1.5em]
\item $\mathcal{R}_{\text{task}}$: \textbf{Rolling task success rate} over a sliding window of the 20 most recent iterations:
$\mathcal{R}_{\text{task}} = \frac{1}{|W|}\sum_{i \in W} \mathbb{1}[\delta_i = 1]$,
where $W$ is the window and $\delta_i$ is the binary success indicator.

\item $\mathcal{R}_{\text{reliab}}$: \textbf{Mean skill reliability} across the current skill library:
$\mathcal{R}_{\text{reliab}} = \frac{1}{|\mathcal{S}|}\sum_{s \in \mathcal{S}} \max(V(s), 0)$.
Since $V(s)$ starts at $0$ for newly created skills and accumulates slowly with successful executions, this component grows on a slower timescale than $\mathcal{R}_{\text{task}}$.

\item $\mathcal{R}_{\text{struct}}$: \textbf{Compositional reuse ratio}, defined as the fraction of skills that are invoked by more than one parent in the network (fan-in $> 1$). A higher value indicates that skills are shared across compositions rather than duplicated.

\item $\mathcal{R}_{\text{cons}}$: \textbf{Refactoring consistency}, measured as the success rate of attempted refactoring operations: $\mathcal{R}_{\text{cons}} = n_{\text{success}} / n_{\text{total}}$ when $n_{\text{total}} > 0$, and $1.0$ when no refactoring has been attempted (no structural changes implies no inconsistency). Both pre-application validation rejections and post-application rollbacks count as failures.
\end{itemize}

The composite objective is a weighted sum:
\begin{equation}
J(\mathcal{N}) = 0.4\,\mathcal{R}_{\text{task}} + 0.3\,\mathcal{R}_{\text{reliab}} + 0.2\,\mathcal{R}_{\text{struct}} + 0.1\,\mathcal{R}_{\text{cons}}.
\end{equation}
The weights reflect a priority ordering: task performance is paramount ($w=0.4$), followed by long-term skill reliability ($w=0.3$), compositional structure ($w=0.2$), and refactoring consistency ($w=0.1$). These weights are hand-selected for interpretability rather than tuned; the qualitative conclusion (stable $J$ under increasing task complexity)is robust across alternative weightings (uniform, inverted, and task-only; see below). $J(\mathcal{N})$ is never explicitly optimized; it serves as a diagnostic that aggregates the health of the skill network across multiple dimensions.

\subsection{Empirical Trajectories}

\begin{figure*}[h]
\centering
\includegraphics[width=\textwidth]{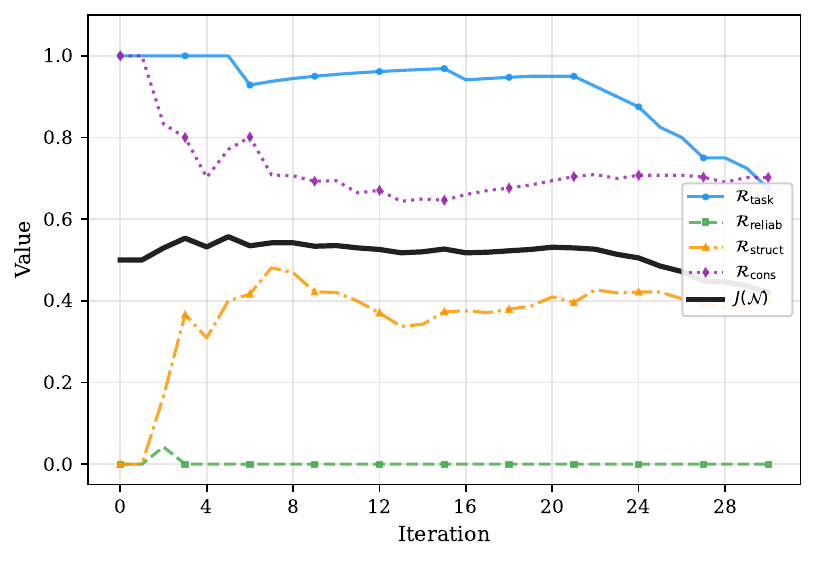}
\caption{Decomposition of $J(\mathcal{N})$ over training iterations (GPT-5-mini, averaged over 2 runs with active refactoring). Each line shows an unweighted component; the black line shows the weighted composite $J(\mathcal{N})$. Despite tasks escalating from wood gathering to diamond tool crafting, $J$ remains stable at $\approx 0.5$: $\mathcal{R}_{\text{task}}$ stays high, $\mathcal{R}_{\text{struct}}$ grows with compositional reuse, and $\mathcal{R}_{\text{cons}}$ stabilizes as refactoring quality improves.}
\label{fig:jn_decomposition}
\end{figure*}

\paragraph{GPT-5-mini: stable learning under increasing complexity.}
The most striking feature is the stability of $J(\mathcal{N}) \approx 0.5$ throughout training (Figure~\ref{fig:jn_decomposition}). This stability is \emph{not} because the tasks are easy: over the iterations, the agent progresses from simple resource gathering (``ensure 4 wood logs'') through multi-step crafting (stone/iron pickaxe, furnace smelting) to deeply compositional objectives (diamond pickaxe, obsidian). The skill graph grows from 1 to $\sim$40 nodes with depth up to 6.

The stability arises from a compensatory dynamic among components. $\mathcal{R}_{\text{task}}$ remains near $1.0$, indicating that PSN successfully solves increasingly difficult tasks without accumulating failures. $\mathcal{R}_{\text{struct}}$ grows steadily from $0$ to $\sim$0.4 as later skills invoke earlier ones, reflecting the network's increasing compositional reuse. The same trajectory on the weaker model (Qwen3) plateaus lower, at $0.15\pm0.07$, consistent with the flatter, less-reused graph reported in Section~\ref{sec:cross_llm}. $\mathcal{R}_{\text{cons}}$ starts high, dips during early refactoring when the network is still small and some proposals fail validation, then stabilizes around $0.7$ as refactoring quality improves with a larger, more structured network. These three components collectively maintain $J$ in a narrow band despite the escalating task difficulty.

This pattern is analogous to a well-regularized neural network whose training loss remains stable as it learns increasingly complex patterns: new representational capacity ($\mathcal{R}_{\text{struct}}$, via skill composition and reuse) is acquired at approximately the same rate that task complexity increases, preventing the objective from degrading.

$\mathcal{R}_{\text{reliab}}$ stays near zero because $V(s)$ accumulates on a slower timescale than task success: newly created skills start at $V(s) = 0$ and require many successful executions to build up Bayesian confidence. 

\paragraph{Summary.}
The stability of $J(\mathcal{N})$ demonstrates that PSN's architectural mechanisms (compositional planning, trace-based optimization, maturity gating, and structural refactoring) enable the agent to absorb increasing task complexity into network structure without degrading task performance. This is a form of ``loss stability'' analogous to well-conditioned training in neural networks: new representational capacity ($\mathcal{R}_{\text{struct}}$) is acquired at approximately the same rate that task difficulty increases, while refactoring consistency ($\mathcal{R}_{\text{cons}}$) stabilizes as the network matures, collectively preventing the composite objective from degrading even as the agent tackles progressively harder compositions.

\paragraph{Weight sensitivity.}
To verify that the stability conclusion does not depend on the specific weight choice, we recompute $J$ under four weighting schemes:

\begin{table}[h]
\small
\centering
\begin{tabular}{lccc}
\toprule
\textbf{Weights} $(w_1,w_2,w_3,w_4)$ & $J_{\text{start}}$ & $J_{\text{end}}$ & Trend \\
\midrule
$(0.4, 0.3, 0.2, 0.1)$ & 0.50 & 0.54 & stable \\
$(0.25, 0.25, 0.25, 0.25)$ & 0.50 & 0.58 & stable \\
$(0.1, 0.2, 0.3, 0.4)$ & 0.50 & 0.62 & stable \\
$(1.0, 0.0, 0.0, 0.0)$ & 1.00 & 0.85 & stable \\
\bottomrule
\end{tabular}
\caption{$J(\mathcal{N})$ under alternative weightings (GPT-5-mini). The stability of $J$ is robust: under all tested weightings, $J$ remains within a narrow range throughout training rather than degrading as task complexity increases.}
\label{tab:weight_sensitivity}
\end{table}

\end{document}